\documentclass[sigconf]{acmart}
\usepackage{times}
\usepackage{graphicx}
\usepackage{amsthm,amsmath}
\usepackage{booktabs}
\usepackage{subfigure}
\usepackage{color}
\usepackage{mathrsfs}
\usepackage{bm}
\usepackage{multirow}
\usepackage{tabularx}
\usepackage[noend]{algpseudocode}
\usepackage{array}
\usepackage{enumitem}
\usepackage{colortbl}
\usepackage{pdftexcmds}
\usepackage{catchfile}
\usepackage{ifluatex}
\usepackage{ifplatform}
\usepackage{caption}
\usepackage{color}
\usepackage{bbm}

\newcolumntype{L}[1]{>{\raggedright\let\newline\\\arraybackslash\hspace{0pt}}m{#1}}
\newcolumntype{C}[1]{>{\centering\let\newline  \\\arraybackslash\hspace{0pt}}m{#1}}
\newcolumntype{R}[1]{>{\raggedleft\let\newline \\\arraybackslash\hspace{0pt}}m{#1}}

\usepackage{booktabs}

\usepackage{algorithm} 
\usepackage{algorithmicx}
\usepackage{algpseudocode}
\def\bA{\textbf{A}}

\def\bH{\textbf{H}}
\def\bW{\textbf{W}}

\def\bh{\textbf{h}}

\usepackage{multirow}
\usepackage[normalem]{ulem}
\useunder{\uline}{\ul}{}

\newcommand\gray{\cellcolor[rgb]{.8,.8,.8}}

\copyrightyear{2022}
\acmYear{2022}
\setcopyright{acmcopyright}
\acmConference[WWW '22] {Proceedings of the ACM Web Conference 2022}{April 25--29, 2022}{Virtual Event, Lyon, France.}
\acmBooktitle{Proceedings of the ACM Web Conference 2022 (WWW '22), April 25--29, 2022, Virtual Event, Lyon, France}
\acmPrice{15.00}
\acmISBN{978-1-4503-9096-5/22/04}
\acmDOI{10.1145/3485447.3512185}

\begin{document}
\fancyhead{}

\title{Designing the Topology of Graph Neural Networks: \\A Novel Feature Fusion Perspective}
\author{Lanning Wei$^{1,2,3}$, 
	Huan Zhao$^{3,*}$, 
	Zhiqiang He$^{1,4,*}$}
\affiliation{
	\institution{$^1$Institute of Computing Technology, Chinese Academy of Sciences \\ $^2$University of Chinese Academy of Sciences  $^3$4Paradigm. Inc., $^4$Lenovo}
	\city{Beijing}
	\country{China}
}
\email{weilanning18z@ict.ac.cn; zhaohuan@4paradigm.com; hezq@lenovo.com}



\begin{abstract}
In recent years, Graph Neural Networks (GNNs) have shown superior performance on diverse real-world applications.
To improve the model capacity, besides designing aggregation operations, GNN topology design is also very important.
In general, there are two mainstream GNN topology design manners.
The first one is to stack aggregation operations to obtain the higher-level features but easily got performance drop as the network goes deeper. 
Secondly, the multiple aggregation operations are utilized in each layer which provides adequate and independent feature extraction stage on local neighbors while are costly to obtain the higher-level information.
To enjoy the benefits while alleviating the corresponding deficiencies of these two manners, 
we learn to design the topology of GNNs in a novel feature fusion perspective which is dubbed F$^2$GNN.
%
To be specific, we provide a feature fusion perspective in designing GNN topology and propose a novel framework to unify the existing topology designs with feature selection and fusion strategies. 
Then we develop a neural architecture search method on top of the unified framework which contains a set of selection and fusion operations in the search space and an improved differentiable search algorithm.
The performance gains on diverse datasets, five homophily and three heterophily ones, demonstrate the effectiveness of F$^2$GNN.
We further conduct experiments to show that F$^2$GNN can improve the model capacity while alleviating the deficiencies of existing GNN topology design manners, especially alleviating the over-smoothing problem, by utilizing different levels of features adaptively.~\footnote{Lanning is a research intern in 4Paradigm. *: Corresponding author. The implementation of F$^2$GNN is available at: \url{https://github.com/AutoML-Research/F2GNN}.}

\end{abstract}

%

\maketitle

\section{Introduction}

\begin{figure}[t]
	\centering
	\includegraphics[width=0.85\linewidth]{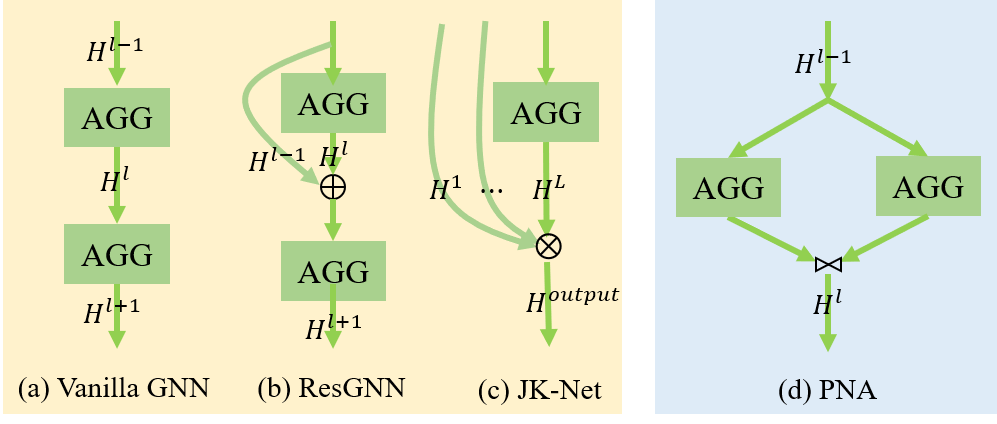}
	\caption{The illustration of two topology design manners. For simplicity, the aggregation operation is denoted by the rectangle. The fusion operation $\oplus$, $\otimes$ and $\Join$ represent the summation, maximum and concatenation, respectively.}
	\label{fig-deeper-wider}
\end{figure}

In recent years, Graph Neural Networks (GNNs) have been widely used due to their promising performance in various graph-based applications~\cite{xu2018powerful,gilmer2017neural,pei2020geom,wei2021pooling,zhang2019autosf}. 
%
In the literature, different GNN models can be built by designing the aggregation operations
\footnote{The aggregation operation in this paper is equivalent to the the message passing layers in GraphGym~\cite{you2020design}.}
and the topology.
To improve the model capacity, diverse aggregation operations~\cite{kipf2016semi,hamilton2017inductive,velivckovic2017graph,xu2018powerful} are designed to aggregate the information from the neighborhood.
%
%
On the other hand, topology design is also important for the model capacity~\cite{cortes2017adanet,xu2018representation,li2019deepgcns}.
One typical topology design manner in GNNs is to stack the aggregation operations.
The higher-order neighborhoods can be accessed based on the stacking manner~\cite{kipf2016semi}, thus, higher-level features can be extracted recursively to increase the model capacity~\cite{li2018deeper}.
%
However, as the network goes deeper, the node representations of connected nodes become indistinguishable,
which is called the over-smoothing problem~\cite{li2018deeper}. 
To address this problem, the identity skip-connection is applied in topology design thus the features of different levels can be utilized to improve the model capacity, e.g., JK-Net~\cite{xu2018representation}, ResGCN~\cite{li2019deepgcns} and DenseGCN~\cite{li2019deepgcns}.
%
Apart from the stacking manner in GNN topology design,
\cite{Gabriele2020PNA,leng2021enhance} use multiple aggregation operations in each layer to extract features independently and fuse these features to enhance the information extraction of local neighbors.
%
%
%
However, multiple aggregations require more resources, hence it is costly to obtain higher-level information when stacking more layers.
To summarize, there are two mainstream GNN topology design manners as shown in Figure~\ref{fig-deeper-wider}
(i) stacking aggregation operations to obtain the higher-level features (methods on the yellow background);
(ii) using multiple aggregation operations to obtain the adequate and independent feature extraction stage on local neighbors.
Both of these two manners can improve model capacity. 
However, the former is easy to get the performance drop due to the over-smoothing problem, and the latter is costly to obtain the higher-level information.
Then when designing a GNN model for a specific task, a natural question arises: \textit{Can we enjoy the benefits while alleviate the corresponding deficiencies on top of these two topology design manners?} 
In that way, we can further improve the model capacity on top of existing GNN models. 

However, it is non-trivial to achieve this since there lacks a systematic approach for the GNN topology design in existing works, and it is unknown how to combine the aforementioned two GNN topology design manners.
In the literature, the topology of a neural network can be represented by its ``computational graph'' where the nodes represent the operations and the directed edges link operations in different layers~\cite{xie2019exploring,valsesia2020don,xie2021understanding,yuan2021differentiable,gu2021dots}. 
In this way, the topology of a neural network can be obtained by designing the links among the nodes in the ``computational graph''.
In existing GNNs, designing the links among operations is equivalent to selecting the features of different levels which proved useful in improving the performance~\cite{valsesia2020don,xu2018representation,you2020design}.
Nevertheless, the fusion strategy designed to make better utilization of the selected features is also indispensable in improving the GNN model capacity~\cite {xu2018representation,zhao2020simplifying,zhao2021search}.
By reviewing extensive human-designed GNNs, we observe that diverse fusion strategies are employed to integrate the features generated by linked operations. 
As the representative GNNs shown in Figure~\ref{fig-deeper-wider}, ResGCN~\cite{li2019deepgcns} adds up the residual features in each layer, JK-Net~\cite{xu2018representation} provides three fusion operations (maximum for example) to integrate the representations of the intermediate layers, and PNA~\cite{Gabriele2020PNA} concatenates the results of multiple aggregation operations.
Therefore, in designing the topology of GNNs, it is important to take the links as well as the fusion strategies into consideration.


On top of the existing GNNs, we design a novel framework to unify the GNN topology designs with feature selection and fusion strategies. Therefore, the topology design target is transformed into the design of these 2 strategies.
As shown in Figure~\ref{fig-search-space}, without loss of generality, the framework is represented as a directed acyclic graph (DAG), which is constructed with an ordered sequence of blocks.
Based on this framework, diverse topology designs including existing works can be modeled by different selection and fusion strategies. 

Then, another challenge is to design adaptive feature selection and fusion strategies to achieve the SOTA performance
given a specific graph-based task, since the preferable GNN topology can significantly differ across different datasets~\cite{you2020design}.
%
Therefore, to address this challenge, we borrow the power of neural architecture search (NAS), which has been successful in designing data-specific CNNs \cite{zoph2016neural,liu2018darts} and GNNs \cite{gao2019graphnas,zhao2021search}, to achieve the adaptive topology design.
To be specific, we firstly propose a novel search space that contains a set of selection and fusion operations, and then develop an improved differentiable search algorithm based on a popular one, i.e., DARTS \cite{liu2018darts}, by addressing the obvious optimization gap induced by two opposite operations in the search space. 
Finally, we extract the optimal strategies based on the unified framework when the searching process terminates, then an adaptive topology design is obtained.

In this paper, we learn to design the topology of GNNs in a novel feature fusion perspective and it can be dubbed F$^2$GNN (Feature Fusion GNN).
Extensive experiments are conducted by integrating the proposed method with predefined and learnable aggregation operations on eight real-world datasets (five homophily and three heterophily), then the performance gains demonstrate the effectiveness of the proposed method.
Furthermore, we conduct experiments to evaluate the advantages of F$^2$GNN in designing the topology of GNNs, from which we can observe that F$^2$GNN can enjoy the benefits and alleviate the deficiencies of existing topology design manners, especially alleviating the over-smoothing problem, by utilizing different levels of features adaptively.

To summarize, the contributions of this work are as follows:
\begin{itemize}
	\item In this paper, we provide a novel feature fusion perspective in designing the GNN topology and propose a novel framework to unify the existing topology designs with feature selection and fusion strategies. It transforms the GNN topology design into the design of this two strategies. 
	\item To obtain the adaptive topology design, we develop a NAS method on top of the unified framework containing a novel search space and an improved differentiable search algorithm.
	\item  Extensive experiments on eight real-world datasets (five homophily and three heterophily) demonstrate that the proposed F$^2$GNN can improve model capacity (performance) while alleviating the deficiencies, especially alleviating the over-smoothing problem.
\end{itemize}

\noindent\textbf{Notations.}
We represent a graph as $\mathcal{G} =(\mathcal{V}, \mathcal{E}) $,where $\mathcal{V}$ and $\mathcal{E}$ represent the node and edge sets. $\textbf{A} \in \mathbb{R}^{|\mathcal{V}| \times |\mathcal{V}|}$ is the adjacency matrix of this graph where $|\mathcal{V}|$ is the node number. The class of node $u$ is represented as $y_u$.
In the proposed framework, the output features of block $i$ are denoted as $\bH^i$, and in other methods, $\bH^i$ represents the output features of layer $i$. $\bH^{output}$ represents the output of a GNN.

\section{Related Work}
%
%
%

\subsection{Topology Designs in Graph Neural Network}

GNNs are built by designing the aggregation operations and topologies. 
One mainstream topology design manner is to stack aggregation operations~\cite{kipf2016semi,hamilton2017inductive,velivckovic2017graph}. The high-level information can be captured to improve the model capacity while easily resulting in the over-smoothing problem in deeper networks~\cite{li2018deeper}.
%
%
To address this problem and improve the model capacity, the identity skip-connections are provided additionally to integrate different levels of features.
JK-Net~\cite{xu2018representation} and DAGNN~\cite{liu2020towards} integrate the features of all the intermediate layers at the end of GNNs; 
ResGCN~\cite{li2019deepgcns} and DenseGCN~\cite{li2019deepgcns} have the same connection schemes as ResNet and DenseNet; GCNII~\cite{chen2020simple} adds up the initial features in each layer.
%
%
%
Apart from the stacking manner, multiple aggregation operations provide adequate and independent feature extraction on the local neighbors to improve the model capacity.
PNA~\cite{Gabriele2020PNA} and HAG~\cite{leng2021enhance} provide multiple aggregation operations in each layer to learn features from local neighbors independently;
%
MixHop~\cite{abu2019mixhop}, IncepGCN~\cite{rong2019dropedge} and InceptionGCN~\cite{kazi2019inceptiongcn}, which are the inception-like methods, provide multiple aggregations in each branch to extract different levels of features independently.
However, multiple aggregations require more resources, hence it is costly to obtain higher-level information.

Considering the benefits and deficiencies of the existing topology designs, we propose a novel method F$^2$GNN in the feature fusion perspective to design the topology of GNNs adaptively.
It can unify the existing topology designs and provide a platform to explore more topology designs, which are more expressive than these human-designed ones.




\subsection{Graph Neural Architecture Search}
\label{sec-related-gnas}
NAS methods use a search algorithm to find the SOTA neural networks automatically in a pre-defined search space and representative methods are~\cite{zoph2016neural,liu2018darts,xie2018snas,li2020sgas}. 
Very recently, researchers tried to automatically design GNNs by NAS. 
These methods mainly focus on designing the aggregation operations on top of the vanilla stacking manner, e.g., GraphNAS~\cite{gao2019graphnas} provides the attention function, attention head number, embedding size, etc. Similar search spaces are also used in~\cite{li2021one}.
Several methods further provide the skip-connection design based on this stacking manner, e.g., SNAE~\cite{zhao2020simplifying} and SNAG~\cite{zhao2021search} are built on JK-Net~\cite{xu2018representation} and learn to select and fuse the features of intermediate layers in the output node, AutoGraph~\cite{li2020autograph} is built on DenseNet~\cite{huang2017densely} and learns to select features in each layer, GraphGym~\cite{you2020design} provides the residual feature selection and fusion strategies in designing GNNs on top of the stacking manner.
Besides these methods, RAN~\cite{valsesia2020don} learns to design the GNN topologies. It uses the computational graph to represent the GNN topology and then designs the links with a randomly-wired graph generator~\cite{xie2019exploring}. However, it lacks the explorations of fusion strategies that can improve feature utilization in GNNs.



%
%
Apart from the search space designs, diverse search algorithms are provided to search the SOTA neural networks from the proposed search space, e.g., incorporating the Reinforcement Learning (RL) into the searching strategy~\cite{zoph2016neural,gao2019graphnas,zhao2020simplifying} or using the Evolutionary Algorithm (EA) directly~\cite{guo2020single,li2020autograph,chen2021graphpas}.
Considering the search efficiency, the differentiable algorithm is proposed to search architectures with gradient descent. It relaxes the discrete search space into continuous and then treats the architecture search problem as a bi-level optimization problem~\cite{liu2018darts,zhao2021search,li2021one,cai2021rethinking}.

More graph neural architecture search methods can be found in~\cite{zhang2021automated,wang2022profiling,ding2021diffmg,wang2021autogel,wei2021pooling,zhang2019autosf,zhang2019interstellar,zhang2021knowledge,qin2021graph}. Compared with existing methods which mainly focus on designing aggregation operations, in this work we explore the additional topology design, which can thus be regarded as an orthogonal and complementary approach to improve the GNN model capacity.

\section{Feature Fusion Framework}
\label{sec-unified-framework}
In this section, we elaborate on the proposed feature fusion framework and show how to translate this framework into diverse GNNs.
\subsection{The Proposed Unified Framework}



In the literature, GNNs can be built by designing the aggregation operation and its topology.
The topology is designed in two major manners: stack aggregation operations to obtain higher-level features or use multiple operations to provide rich and independent local feature extractions.
By reviewing existing human-designed GNNs, we observe that features of different levels are selected due to diverse selection strategies, and it leads to the utilization of diverse fusion strategies to integrate these features~\cite {li2019deepgcns,xu2018representation,abu2019mixhop,Gabriele2020PNA}. 
In other words, the feature selection and fusion strategies lead to the key difference of topology designs in GNNs. Therefore, the topology designs utilized in existing methods can be unified with these two strategies.

\begin{figure}[t]
	\centering
	\includegraphics[width=0.9\linewidth]{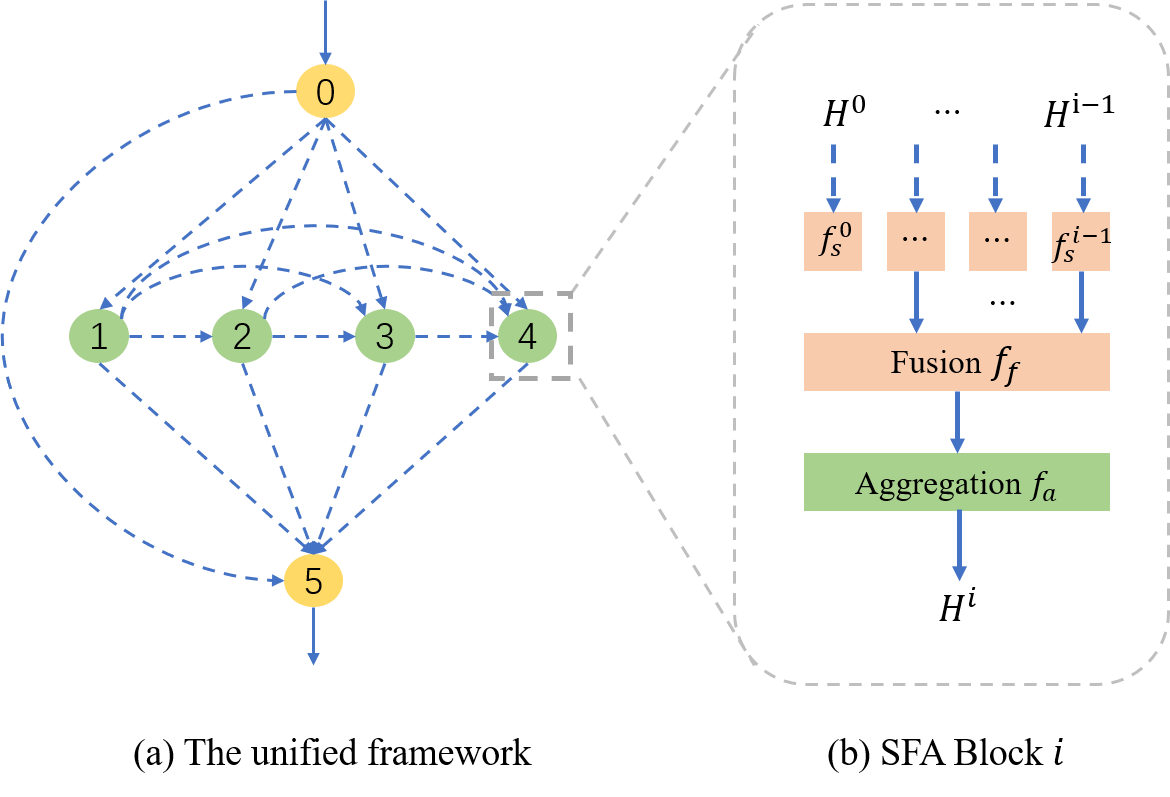}
	\caption{(a) The proposed framework consists of an ordered sequence of one input block, $N$ SFA blocks (four as an example here) and one output block. 
		(b) For the $i$-th SFA block, we have $i$ selection operations $f_s$ and one fusion operation $f_f$ to utilize the features generated by $i$ predecessors. Then one aggregation operation $f_a$ is followed to aggregated messages from the neighborhood.}
	\label{fig-search-space}
\end{figure}

Based on this motivation, we propose a feature fusion framework to unify these two topology design manners. As shown in Figure~\ref{fig-search-space}, without loss of generality, the framework is represented as a DAG which is constructed with an ordered sequence of one input block, $N$ SFA blocks ($N=4$ for example; SFA: \underline{s}election, \underline{f}usion and \underline{a}ggregation), and one output block.
The input block only contains a simple pre-process operation, i.e., Multilayer Perceptron (MLP) in this paper, that supports the subsequent blocks.
%
%
%
The SFA block contains the selection, fusion and aggregation operations. 
For the $i$-th SFA block as shown in Figure~\ref{fig-search-space}(b), there exists $i$ predecessors thus $i$ feature selection operations $f_s$ are utilized to select the features generated by the previous blocks.
One fusion operation $f_f$ is used to fuse these selected features, and one aggregation operation $f_a$ is followed to aggregate messages from the neighborhood. 
Therefore, the high-level features can be generated by $\bH^i = f_a(f_f(\{f_s^0(\bH^0), \cdots, f_s^{i-1}(\bH^{i-1})\}))$.
%
In the output block, a 2-layer MLP that serves as the post-process operation is provided after the $N+1$ selection operations and one fusion operation.
On top of the unified framework, the topology design is transformed into the design of selection and fusion strategies. 
Compared with existing methods that focus on designing aggregation operations, our framework provides a platform to explore the GNN topology designs which is more expressive than existing methods. 
\begin{figure*}[t]
	\label{fig-approximation}
	\begin{minipage}{0.25\linewidth}
		\centering
		\footnotesize
		\begin{tabular}{c|l}
			\hline
			Method & Equation \\ \hline
			Vanilla GNN  & $\bH^{l+1} = f_a(\bH^l)$         \\ \hline
			ResGCN~\cite{li2019deepgcns}    & $\bH^{l+1} = f_a(\bH^l + \bH^{l-1})$         \\ \hline
			JK-Net~\cite{xu2018representation}     & $\bH^{output} = f_f( \bH^1,\cdots, \bH^{L}) $         \\ \hline
			GCNII~\cite{chen2020simple}  & $\bH^{l+1} =  \alpha \bH^0 + (1-\alpha)f_a(\bH^l)$         \\ \hline
			PNA~\cite{Gabriele2020PNA}    & $\bH^{l+1} = \mathop{\parallel}\limits_{i \in M} f_a(\bH^l) $         \\ \hline\			MixHop~\cite{abu2019mixhop} & $\bH^{l+1} = \mathop{\parallel}\limits_{i \in P} \bA^i\bH^l$         \\ \hline
		\end{tabular}
	\end{minipage}
	\hfill
	\begin{minipage}{0.7\linewidth}
		\centering
		\includegraphics[width=0.95\linewidth]{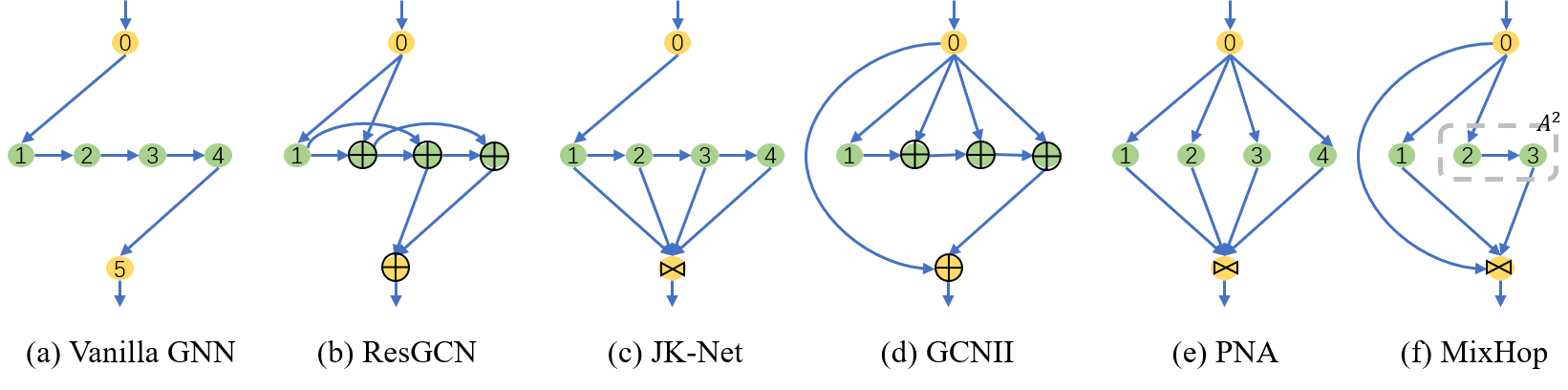}
	\end{minipage}
	\caption{The formulation and illustrations of the closely related methods. $\oplus$ denotes the summation operation and $\Join$ denotes the concatenation operation.}
\end{figure*}

\subsection{Translating the framework into diverse GNNs}

The feature fusion framework can unify the existing topology designs thus we can translate the framework into diverse GNNs.
The formulation and illustrations of the most closely related methods are shown in Figure~3. For simplicity, the feature transformation is ignored.
Vanilla GNNs~\cite{kipf2016semi,hamilton2017inductive,velivckovic2017graph,xu2018powerful} are constructed by stacking aggregation operations and these methods can be approximated as the single path in our framework.
ResGCN~\cite{li2019deepgcns} and JK-Net~\cite{xu2018representation} provide extra identity skip-connections to utilize the different levels of features, thus diverse selection and fusion operations are utilized as illustrated in Figure~3(b) and (c), respectively. 
GCNII~\cite{chen2020simple} uses the initial residual connections to address the over-smoothing problem, and it can be approximated as in Figure~3(d). 
PNA~\cite{Gabriele2020PNA} uses $M$ aggregation operations in each layer, and it can be approximated by $M$ branches in this framework.
MixHop~\cite{abu2019mixhop} concatenates the features based on the power of adjacency matrix $\bA^i$. $P$ is the set of the adjacency powers, and this method can be approximated as in Figure~3(f).
%
For the graph neural architecture search methods which focus on design aggregation operations~\cite{gao2019graphnas,li2021one} or provide additional skip-connection learning~\cite{zhao2020simplifying,zhao2021search} on top of the stacking manner, they can be treated as the variants of vanilla and JK-Net, respectively.
Thus, we can see the advantage of the proposed framework, which can provide a systematic and unified view of existing GNN topology designs. Next, we show how to obtain the data-specific GNN topology design on top of this unified framework.

\section{Design GNNs with the framework}

On top of the unified framework, generating one adaptive topology design is a non-trivial task since it is still unknown how to design the optimal feature selection and fusion strategies given a specific task.
In this paper, we adopt the NAS method to address this challenge.
As the commonly-recognized paradigm of NAS, the search space which contains the selection and fusion operation sets is provided. 
Then the topology design can be decomposed into the decision of operations in the search space.
To demonstrate the versatility of our method, we further provide the aggregation operation set thus we can design GNNs with topologies and aggregation operations simultaneously.
The differentiable search algorithm~\cite{liu2018darts} is widely used considering the efficiency and effectiveness. Nevertheless, we observe the performance gap in applying this algorithm due to the two opposite selection operations, i.e., \texttt{ZERO} and \texttt{IDENTITY}. Thus, we develop a new algorithm to alleviate this problem.

\subsection{Search Space}
\label{sec-search-space}


One GNN can be built by designing the aggregation operations and the topology. 
The aggregation operations are widely explored in existing methods. In general, GNNs use the same aggregation operation in each layer. \cite{gao2019graphnas,zhao2020simplifying,zhao2021search} learn to design the layer-wise aggregations with the help of the NAS methods as introduced in Section~\ref{sec-related-gnas}.
Therefore, in this paper, we design the topology designs on these two aggregation variants.

\subsubsection{Predefined aggregation operation}
The predefined aggregation is utilized in each SFA block and then we need to design the selection and fusion operations in the framework with the help of NAS. After the search terminates, we can obtain the designed GNN with the given aggregation operation. We provide 2 aggregation operations GraphSAGE and GAT thus this method can be dubbed F$^2$SAGE and F$^2$GAT, respectively. Other aggregation operations like GCN and GIN can be trivially integrated with the proposed framework.
Without loss of generality, we provide a set of candidate selection operations $\mathcal{O}_s$ and fusion operations $\mathcal{O}_f$ in the search space as shown in Table~\ref{tb-search-space}. 



\noindent\textbf{Selection Operations.}
For the selection operation, there are only ``selected'' and ``not selected'' stages for each feature in existing methods. Thus, we provide two operations \texttt{IDENTITY} and \texttt{ZERO} to select features, which can be represented as $f(\bh)=\bh$ and $f(\bh)=\textbf{0}\cdot\bh$, respectively.

\noindent\textbf{Fusion Operations.} In SFA block and the output block, one fusion operation is needed to fuse the selected features. Based on the literature, we provide six fusion operations to fuse these features with the summation, average, maximum, concatenation, LSTM cell and attention mechanism, which are denoted as \texttt{SUM}, \texttt{MEAN}, \texttt{MAX}, \texttt{CONCAT}, \texttt{LSTM} and \texttt{ATT}, respectively.

\subsubsection{Learnable aggregation operation}
Compared with the existing graph neural architecture search methods which focus on designing the aggregation operations, we provide the extra aggregation operation set $\mathcal{O}_a$ thus we can design GNNs with topologies and aggregation operations simultaneously. This method is dubbed F$^2$GNN.
\noindent\textbf{Aggregation Operations.} 
Four widely used aggregation operations are used in this paper : GCN~\cite{kipf2016semi}, GAT~\cite{velivckovic2017graph}, GraphSAGE~\cite{hamilton2017inductive} and GIN~\cite{xu2018powerful}, which denoted as \texttt{GCN}, \texttt{GAT}, \texttt{SAGE} and \texttt{GIN}, respectively. 
In this paper, we focus on designing the topology of GNNs thus only four aggregations are provided. More operations in existing methods can be trivially added if needed.



\begin{table}[]

	\caption{The operations used in our search space.}
	\begin{tabular}{l|l}
		\hline
		 & Operations            \\ \hline
		Selection $\mathcal{O}_s$ & \texttt{ZERO}, \texttt{IDENTITY} \\ \hline
		Fusion $\mathcal{O}_f$&\texttt{SUM}, \texttt{MEAN}, \texttt{MAX}, \texttt{CONCAT}, \texttt{LSTM}, \texttt{ATT}  \\ \hline
		Aggregation $\mathcal{O}_a$ & \texttt{GCN}, \texttt{GAT}, \texttt{SAGE}, \texttt{GIN}\\ \hline
	\end{tabular}
\label{tb-search-space}
\end{table}                                                                                                                     
\subsection{The Improved Search Algorithm}
\label{sec-search-algo}

Based on the proposed framework and the search space, the search algorithm is used to search operations from the corresponding operation set. 
Considering the efficiency and effectiveness which have been demonstrated in \cite{liu2018darts,zhao2021search}, without loss of generality, the differentiable search algorithm is employed.

\noindent\textbf{Preliminary: Differentiable Architecture Search.}
A supernet is defined to subsume all models on top of the unified framework and the search space, and it is achieved by mixing the candidate operations~\cite{guo2020single,liu2018darts}. The results of the mixed operation can be calculated by a weighted summation of all candidate operations which denoted as $\bar{o}(x)=\sum\nolimits_{k=1}^{\left|\mathcal{O}\right|} c_ko_k(x)$, where $c_k \in (0,1)$ is the weight of $k$-th candidate operation $o_k(\cdot)$. 
In general, the operation weight $c_k$ is generated by one relaxation function $c_k=  \frac{\exp(\alpha_k)}{\sum\nolimits_{i=1}^{\left|\mathcal{O}\right|} \exp(\alpha_i)}$ and $\alpha_k$ is the corresponding learnable supernet parameter for $c_k$~\cite{liu2018darts}.
%
%
Based on the relaxation function, the discrete selection of operations is relaxed into continuous and we can generate the final results in the output block step by step as shown in Figure~\ref{fig-search-space}. Thus the supernet can be optimized with gradient descent which can accelerate the search process in orders of magnitude.
After finishing the search process, we preserve the operations with the largest weights in each mixed operation, from which we obtain the searched GNN.


\noindent\textbf{The optimization gap in feature fusion.}
We optimize the supernet in the search process and then derive the GNN after the search is finished. 
However, it is difficult to generate the best childnet from the supernet since we optimize the supernet in the search process and only select the childnet in reality, which is called optimization gap~\cite{chen2021progressive,xie2021weight} and the evaluation results can be found in Section~\ref{sec-ablation-gap}. 
The performance drop caused by the optimization gap is extremely obvious in the our method since we provide two opposite operations in the selection operation set. In the following, we briefly explain this problem.

For the features generated by block $i$, the results of mixed selection operation in block $j$ ($j>i$) can be represented as the weighted summation of \texttt{ZERO} and \texttt{IDENTITY} operations as shown in    
\begin{align}
	\label{eq-block-output}
	\bar{o}^{ij}(\textbf{x}_i)=\sum\nolimits_{k=1}^{\left|\mathcal{O}_s\right|} c_k^{ij}o_k^{ij}(\textbf{x}_i) = c_1^{ij}\textbf{0} + c_2^{ij}\textbf{x}_i=c_2^{ij}\textbf{x}_i.
\end{align} 
Then the results of mixed fusion operation in block $j$ can be generated by
\begin{align}
	\label{eq-combination}
	\bar{o}^j(\textbf{x}) = \sum\nolimits_{k=1}^{\left|\mathcal{O}_f\right|} c_k^{j}o_k^j(\textbf{x}) = \sum\nolimits_{k=1}^{\left|\mathcal{O}_f\right|} c_k^{j}f_k^j(\{\bar{o}^{ij}(\textbf{x}_i)|i=0,\cdots,j-1\}),
\end{align}
where $f_k^j$ is the $k$-th candicate fusion operation in block $j$. 

When the weight of \texttt{ZERO} operation $c_1$ is larger than the weight of \texttt{IDENTITY} operation $c_2$, in the childnet, the \texttt{ZERO} operation should be chosen and one zero tensor results in Eq.~\eqref{eq-block-output} are expected. Furthermore, one zero tensor results will be obtained in Eq.~\eqref{eq-combination} if no feature is selected in this block. 
However, in the supernet, the mixed operation results $c_2^{ij}\textbf{x}_i$ in Eq.~\eqref{eq-block-output} will be generated. 
That is, in this case, one zero tensor result is expected while we got $c_2^{ij}\textbf{x}_i$ in reality in each mixed selection operation. The \texttt{IDENTITY} operation has a large influence in Eq.~\eqref{eq-block-output} when \texttt{ZERO} is selected, and the influence will accumulate along with the feature selection operation in the framework.
Therefore, the gap between the supernet results and the childnet results in our framework is extremely obvious due to these two opposite selection operations, and we cannot derive the best childnet from the supernet due to this gap as the evaluation in Section~\ref{sec-ablation-gap}. 

%

%

\noindent\textbf{Improved search with the usage of temperature.} Considering the influence of \texttt{IDENTITY} operation, we add a temperature in Softmax function as  $c_k=  \frac{\exp(\alpha_k/\lambda)}{\sum\nolimits_{i=1}^{\left|\mathcal{O}\right|} \exp(\alpha_i/\lambda)}$.
Thus, with a small temperature $\lambda$, the operation weight vector \textbf{c} close to a one-hot vector, and the results of mixed selection operation in Eq.~\eqref{eq-block-output} close to a zero tensor when the \texttt{ZERO} operation is selected. That is, the \texttt{IDENTITY} operation will have a smaller influence on the selection results when the \texttt{ZERO} operation is chosen, and the optimization gap in our method can be alleviated. Similar solutions can be found in ~\cite{xie2018snas,chen2019progressive}.
In this paper, we set $\lambda=0.001$ and the influence of different temperatures will be shown in Section~\ref{sec-ablation-gap}.

\noindent\textbf{Deriving process.}
The architecture searching task is treated as the bi-level optimization problem.
Our method is optimized with gradient descent introduced in~\cite{liu2018darts,zhao2021search}. More details can be found in Appendix~\ref{appendix-alg}.
After finishing the search process, we preserve the operation with the largest weight in each mixed operation, from which we obtain the searched architecture.

\section{Experiments}


\subsection{Experimental Settings}
\label{sec-exp-set}
\noindent\textbf{Datasets.} 
Existing GNNs assume strong homophily where neighbors are in the same class. The homophily ratio can be calculated by $h=\frac{\left| \{(u,v):(u,v) \in \mathcal{E} \wedge y_u=y_v\}\right|}{\left| \mathcal{E} \right|}$ which is the fraction of edges in a graph which connect nodes that have the same class label. 
As shown in Table \ref{tb-dataset}, five widely used homophily datasets (higher homophily ratio $h$) and three heterophily datasets (lower homophily ratio $h$) are selected to evaluate the performance of our method. More introductions about these datasets can be found in Appendix~\ref{appendix-dataset}. 
The comparisons between the F$^2$GNN and other heterophily methods~\cite{zhu2020beyond,du2021gbk} will be considered in future work.



\begin{table}[]
	\centering
	\small
	\caption{Statistics of the eight datasets in our experiments.}
\begin{tabular}{c|c|c|c|c|c}
	\hline
	Datasets                          & \#Nodes & \#Edges & \#Features & \#Classes & $h$    \\ \hline
	Cora~\cite{sen2008collective}     & 2,708   & 5,278   & 1,433      & 7         & 0.81 \\ 
	Computers~\cite{mcauley2015image} & 13,381  & 245,778 & 767        & 10        & 0.78 \\
	DBLP~\cite{bojchevski2017deep}    & 17,716  & 105,734 & 1,639      & 4         & 0.83 \\
	PubMed~\cite{sen2008collective}   & 19,717  & 44,324  & 500        & 3         & 0.80 \\ 
	Physics~\cite{shchur2018pitfalls} & 34,493  & 495,924 & 8,415      & 5         & 0.93 \\ \hline
	Wisconsin~\cite{pei2020geom}      & 251     & 466     & 1,703      & 5         & 0.21 \\
	Actor~\cite{pei2020geom}          & 7,600   & 30,019  & 932        & 5         & 0.22 \\ 
	Flickr~\cite{zeng2019graphsaint}  & 89,250  & 899,756 & 500        & 7         & 0.32 \\ 
	\hline
\end{tabular}
\label{tb-dataset}
\end{table}

\noindent\textbf{Baselines.}
On top of the predefined aggregations, we provide nine GNNs constructed with different topologies.
%
%
(a) GNNs are constructed by stacking two and four aggregation operations;
(b) based on the stacking manner, we construct 4-layers GNNs on top of three commonly used topology designs ResGCN~\cite{li2019deepgcns}, DenseGCN~\cite{li2018deeper} and JK-Net~\cite{xu2018representation}. They are denoted as RES, DENSE and JK, respectively;
(c) 4-layer GNNII is constructed based on the topology shown in Figure 3(d);
(d) the topology of 1-layer PNA and MixHop are shown in Figure 2(e) and Figure 2(f), respectively. The PNA and MixHop baselines in our experiment are constructed by stacking two layers.
(e) the topology designs are constructed by selecting operations from the search space randomly, which is denoted as Random in our experiments.

%
%
Compared with F$^2$GNN which designs the topology and aggregation operations simultaneously, we provide three graph neural architecture search baselines: (a) an RL based method SNAG~\cite{zhao2020simplifying}, (b) a differentiable method SANE~\cite{zhao2021search}, and (c) a random search algorithm that uses the same search space as F$^2$GNN. 

More details of these baselines are provided in Appendix~\ref{appendix-baseline}.

\begin{table*}[]
	\centering
	\footnotesize
		\caption{Performance comparisons of our method and all baselines. We report the average test accuracy and the standard deviation with 10 splits. ``L2'' and ``L4'' mean the number of layers of the base GNN architecture, respectively. The best result in each group is highlighted in gray, and the second best one is underlined. 
		The group accuracy rank and the overall accuracy rank of each method are calculated on each dataset. The average rank on all datasets is provided. The Top-2 methods in each group and the Top-3 methods in this table are highlighted in gray.}
	\label{tb-performance-agg}
\begin{tabular}{c|c|ccccc|ccc|cc}
\toprule
	Aggregation                & Topology      & Cora              & DBLP              & PubMed            & Computers         & Physics           & Actor             & Wisconsin         & Flickr            & \begin{tabular}[c]{@{}c@{}}Avg. Rank\\ (Group)\end{tabular}          & \begin{tabular}[c]{@{}c@{}}Avg. Rank\\ (All)\end{tabular}          \\ \hline
	\multirow{10}{*}{SAGE}     & Stacking (L2) & 86.09(0.50)       & 83.58(0.33)       & 88.96(0.29)       & 91.14(0.30)       & 96.42(0.11)       & 34.78(1.10)       & 79.61(5.56)       & 51.21(0.71)       & 6.63       & 15.00      \\
	& Stacking (L4) & 85.68(0.61)       & 83.83(0.32)       & 88.23(0.28)       & 90.52(0.42)       & 95.97(0.14)       & 34.61(1.08)       & 60.39(10.77)      & 53.07(0.50)       & 8.25       & 17.00      \\
	& RES (L4)      & 85.66(0.52)       & 83.39(0.30)       & 88.99(0.25)       & {\ul 91.51(0.18)} & 96.31(0.17)       & 35.16(0.94)       & 76.47(5.26)       & {\ul 53.72(0.27)} & 5.25       & 13.13      \\
	& DENSE (L4)    & 86.68(0.59)       & 83.30(0.73)       & {\ul 89.42(0.27)} & 90.74(0.51)       & 96.48(0.14)       & 34.78(0.60)       & 77.06(6.01)       & 53.17(0.19)       & 4.50       & 12.75      \\
	& JK (L4)       & 86.47(0.60)       & 83.94(0.62)       & 89.21(0.29)       & 91.21(0.30)       & {\ul 96.56(0.05)} & {\ul 36.53(0.92)} & 81.96(4.71)       & 52.41(0.33)       & 4.75       & 10.38      \\
	& GNNII (L4)    & 85.83(0.42)       & {\ul 84.46(0.45)} & 89.21(0.24)       & 91.38(0.27)       & 96.45(0.15)       & 35.70(1.11)       & 81.57(4.13)       & 52.24(0.29)       & 4.50       & 11.50      \\
	& PNA (L2)      & 84.29(0.67)       & 82.76(0.42)       & 89.25(0.26)       & 90.67(0.42)       & 96.32(0.10)        & 33.89(2.68)       & 75.29(6.46)       & 52.09(0.73)       & 8.88       & 17.75      \\
	& MixHop (L2)   & 84.81(0.95)       & 82.65(0.65)       & 89.25(0.28)       & 88.56(1.61)       & 96.11(0.17)       & 35.19(0.62)       & 81.57(2.51)       & 51.75(0.59)       & 6.75       & 17.75      \\
	& Random        & {\ul 86.75(0.29)} & 83.60(0.29)       & 89.21(0.04)       & 91.30(0.19)       & 96.46(0.03)       & 36.30(0.58)       & {\ul 85.10(5.63)} & \gray 54.10(0.15) & \gray 3.50 & 8.75       \\
	& F$^2$SAGE     & \gray 87.72(0.26) & \gray 84.81(0.06) & \gray 89.73(0.26) & \gray 91.81(0.26) & \gray 96.72(0.01) & \gray 36.61(1.00) & \gray 85.88(1.92) & 53.66(0.16)       & \gray 2.00 & \gray 4.38 \\ \hline
	\multirow{10}{*}{GAT}      & Stacking (L2) & 85.92(0.72)       & {\ul 84.34(0.26)} & 87.56(0.23)       & 91.49(0.21)       & 95.76(0.16)       & 29.28(1.02)       & 53.73(7.24)       & {\ul 53.83(0.28)} & 5.25       & 14.25      \\
	& Stacking (L4) & 86.16(0.55)       & 84.29(0.41)       & 85.73(0.34)       & 89.08(0.43)       & 93.47(3.93)       & 26.45(1.00)       & 45.29(5.65)       & 50.34(2.68)       & 8.25       & 19.88      \\
	& RES (L4)      & 84.66(0.92)       & 84.11(0.34)       & 87.56(0.44)       & 90.84(0.49)       & 95.67(0.28)       & 28.98(0.36)       & 48.82(3.77)       & 53.63(0.24)       & 7.50       & 18.50      \\
	& DENSE (L4)    & 85.31(0.86)       & 83.43(0.37)       & 88.67(0.19)       & 91.30(0.37)       & 96.16(0.06)       & 31.78(1.03)       & 53.33(7.73)       & 53.61(0.26)       & 6.25       & 16.38      \\
	& JK (L4)       & {\ul 86.55(0.46)} & 83.73(0.35)       & {\ul 89.71(0.16)} & 91.80(0.23)       & 96.80(0.09)       & 35.43(0.88)       & 84.51(5.58)       & 53.02(0.29)       & \gray 3.88 & 8.75       \\
	& GNNII (L4)    & 85.40(1.06)       & 83.83(0.33)       & 88.44(0.25)       & {\ul 91.91(0.11)} & 96.14(0.15)       & 30.29(0.78)       & 55.29(6.25)       & 53.03(0.29)       & 5.38       & 15.00      \\
	& PNA (L2)      & 85.06(0.72)       & 83.46(0.47)       & 87.18(0.30)       & 90.84(0.24)       & 95.85(0.18)       & 28.56(0.82)       & 49.22(5.91)       & \gray 54.02(0.33) & 7.38       & 18.25      \\
	& MixHop (L2)   & 85.38(1.04)       & 82.50(0.34)       & 88.91(0.19)       & 91.27(0.37)       & 96.46(0.21)       & 35.70(0.90)       & 81.57(4.40)       & 53.67(0.30)       & 5.13       & 13.25      \\
	& Random        & 85.73(0.06)       & 83.60(0.19)       & 88.86(0.18)       & 91.76(0.14)       & {\ul 96.84(0.09)} & {\ul 36.07(0.83)} & {\ul 86.08(4.15)} & 52.43(0.29)       & 4.38       & 10.38      \\
	& F$^2$GAT      & \gray 88.31(0.12) & \gray 84.76(0.04) & \gray 90.38(0.14) & \gray 92.04(0.17) & \gray 97.10(0.03)  & \gray 36.65(1.13) & \gray 87.06(4.13) & 53.45(0.19)       & \gray 1.63 & \gray 3.13 \\ \hline
	\multirow{4}{*}{Learnable} & SNAG (L4)     & 84.99(1.04)       & 84.29(0.15)       & 87.93(0.16)       & 85.98(0.72)       & 96.18(0.11)       & 28.13(0.74)       & 43.92(4.65)       & 53.50(0.31)       & 4.00       & 18.63      \\
	& SANE (L4)     & 86.40(0.38)       & 84.58(0.13)       & 89.34(0.31)       & 91.02(0.21)       & {\ul 96.80(0.06)} & {\ul 36.77(1.15)} & {\ul 86.47(3.09)} & 53.92(0.14)       & 2.63       & 6.38       \\
	& Random        & {\ul 86.99(0.60)} & {\ul 84.62(0.15)} & {\ul 89.37(0.26)} & {\ul 91.03(0.20)} & 96.72(0.04)       & 36.29(1.52)       & 85.49(4.31)       & \gray 54.33(0.11) & \gray 2.25 & 6.13       \\
	& F$^2$GNN      & \gray 87.42(0.42) & \gray 84.95(0.15) & \gray 89.79(0.20) & \gray 91.42(0.26) & \gray 96.92(0.06) & \gray 37.08(1.00) & \gray 88.24(3.72) & {\ul 53.96(0.20)} & \gray 1.13 & \gray 2.75 \\ \bottomrule
\end{tabular}
\end{table*}

\noindent\textbf{Implementation details.}
%
For Cora, DBLP, Computers, PubMed and Physics, we split the dataset with 60\% for training, 20\% for validation and test each considering supernet training and evaluation. 
For Wisconsin and Actor, we adopt the 10 random splits used in~\cite{pei2020geom,zhu2020beyond}(48\%/32\%/20\% of nodes per class for train/validation/test).
For Flickr dataset, we adopt the split in~\cite{zeng2019graphsaint}(50\%/25\%/25\% of nodes per class for train/validation/test).
For all NAS methods (Random baselines, SNAG, SANE and our method), we search a GNN with the corresponding search space. Then all the searched GNNs and the human-designed baselines are tuned individually with hyperparameters like embedding size, learning rate, dropout, etc. 
With the searched hyperparameters, we report the average test accuracy and the standard deviation on 10 repeated results.
%
More details about the implementation stage are shown in Appendix~\ref{appendix-implementation}.
\subsection{Performance Comparisons}

The results are given in Table \ref{tb-performance-agg}. 
Firstly, based on two predefined aggregation operations, there is no absolute winner among eight human-designed GNNs constructed with existing topology designs.
Among them, with the utilization of different feature selection and fusion strategies, six baselines have a better performance than two stacking baselines in two groups in general. 
The performance gain demonstrates the importance of feature utilization in improving the model capacity.
Secondly, by designing GNN topologies with adaptive feature selection and fusion strategies, the proposed method can achieve the top rank on two predefined aggregations. In particular, the SOTA performance is achieved on seven out of eight datasets.
%
%
Thirdly, with the same search space, the Random baselines also achieve considerable performance gains on all these datasets, which demonstrates the effectiveness of the unified framework in Section \ref{sec-unified-framework}. 
Nevertheless, the Random baseline is outperformed by the proposed method, which indicates the usefulness of the improved search algorithm in designing the topology of GNNs.

\begin{figure}[t]
	\centering
	\includegraphics[width=0.8\linewidth]{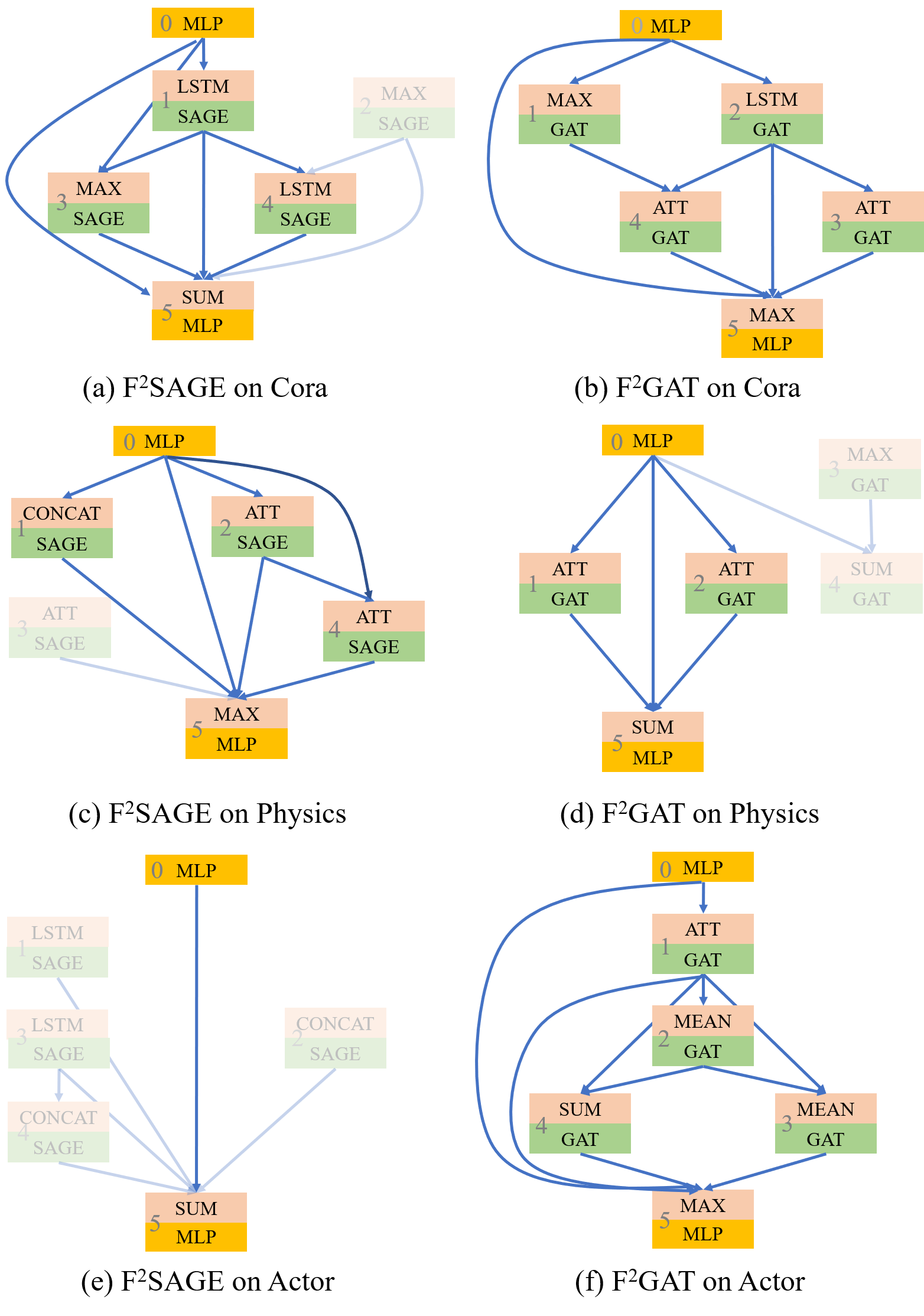}
	\caption{The searched topologies on Cora, Physics and Actor datasets with two human-designed aggregation operations: GraphSAGE and GAT. The index of each block in the framework is annotated and the dark blocks indicate they are not used in the searched GNNs.}
	\label{fig-searched_arch}
\end{figure}

On the other hand, looking at the results of the learnable aggregation operations, SANE and SNAG only focus on selecting features in the output block based on a stacking manner (A sketch of the topology is given in Figure \ref{fig-sane-framework} in Appendix~\ref{appendix-baseline}.). These methods have limited topology design space and can be treated as special instance of F$^2$GNN. 
Compared with these two methods, F$^2$GNN and Random methods design GNNs on the proposed unified framework and achieve higher performance. Thus it demonstrates the effectiveness of the adaptive topology design, i.e., feature selection and fusion strategies based on the unified framework. 
Besides, \cite{liu2018darts,zhao2021search} have shown the efficiency advantage of the differentiable methods (our method and SANE) over those RL and randomly based methods. We also obtained the two orders of magnitude less search cost in our experiments (The details of the search cost comparison are shown in Section~\ref{sec-ablation-gap}.).
These 2 differentiable methods stay at the top of the rank list (F$^2$GNN ranks 2.75 and SANE ranks 6.38), which indicates the power of differentiable search algorithms in designing GNNs. 

\noindent\textbf{Searched topologies.}
We visualize the searched topologies on Cora, Physics and Actor datasets with different aggregation operations in Figure~\ref{fig-searched_arch}.
We emphasize on several interesting observations in the following:

\noindent$\bullet$ The topologies are different for different aggregation operations and datasets. The performance gain and the top ranking demonstrate the necessity of designing data-specific GNN topologies.

\noindent$\bullet$ The initial features generated by the input block are utilized in the output block in almost all GNNs. These features contain more information about the node itself which are important for node representation learning as mentioned in~\cite{chen2020simple,zhu2020beyond}.

\noindent$\bullet$ We can benefit from the multiple aggregation design manner which provides adequate and independent local feature extractions, e.g., two aggregation operations are selected in the second layer on Cora and the first layer on Physics.

\noindent$\bullet$ On Actor, the representative heterophily dataset, we obtained an MLP network based on F$^2$SAGE. This topology design is consistent with H$_2$GCN ~\cite{zhu2020beyond} which shows that the graph structure is not always useful for the final performance, and it further demonstrates the effectiveness and the versatility of our method. More searched topologies are shown in Appendix~\ref{appendix-searched} and more results about heterophily datasets are given in Appendix~\ref{appendix-hetero}.

\subsection{Advantages of the Adaptive Topology Design}
\label{sec-exp-advantage}
In designing the topology of GNNs, stacking aggregation operations devoted to obtaining higher-level features but easily got the over-smoothing problem as the network goes deeper, and the multiple aggregation operations provide adequate and independent feature extraction stage on local neighbors while are costly to obtain the higher-level information.
The performance gains in Table~\ref{tb-performance-agg} and the searched topology designs in Figure~\ref{fig-searched_arch} indicate the effectiveness of designing the topology of GNNs with two design manners.
With the proposed F$^2$GNN which can design topology with the NAS method adaptively, we show the advantages of our method in alleviating the deficiencies of the existing two topology design manners.

\begin{figure}[t]
	\centering
	\includegraphics[width=0.85\linewidth]{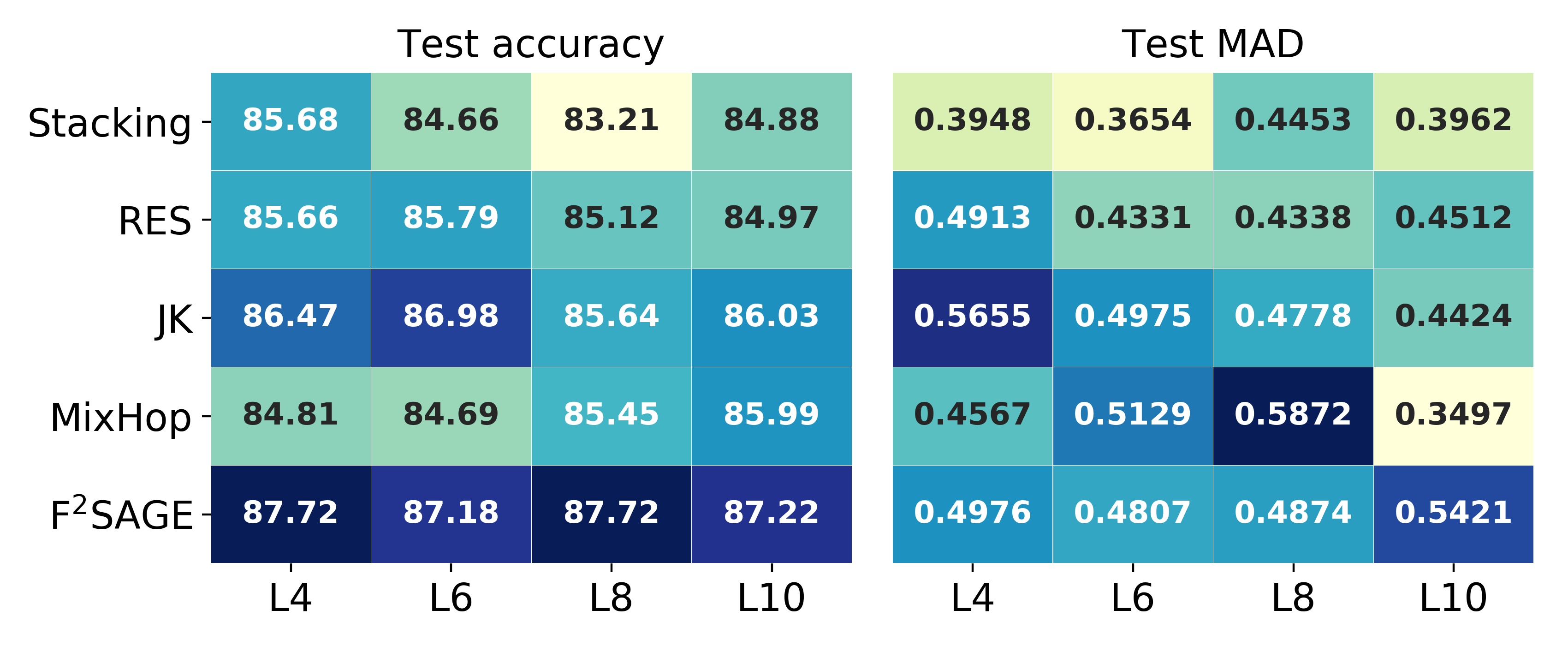}
	\caption{Comparisons of test accuracy and the MAD value on the Cora dataset. The predefined aggregation operation is the GraphSAGE.
		``L4'' represents the 4 layer baseline ($N=4$ in our method), and so on. Darker colors mean larger values. }
	\label{fig-mad-layer}
\end{figure}

\noindent\textbf{Alleviating the over-smoothing problem.} 
As the network goes deeper, the node representations become indistinguishable and easily got performance drop, which is called over-smoothing~\cite{li2018deeper}. 
%
%
MAD (Metric for Smoothness)~\cite{chen2020measuring} is used to measure the smoothness of the features. 
In Figure~\ref{fig-mad-layer}, we show the comparisons of test accuracy and MAD value on the Cora dataset on the conditions of different layers and SFA blocks. 
For comparisons, on top of the GraphSAGE, we provide three topologies that have been proved helpful in alleviating the over-smoothing problem, and more results can be found in Appendix~\ref{appendix-oversmoothing}.
Other methods~\cite{rong2019dropedge,feng2020graph,chen2020measuring} which can alleviate this problem will be left into future work. 
In Figure~\ref{fig-mad-layer}, RES, JK and MixHop can achieve stable performance and higher MAD values compared with the stacking baseline. It demonstrates the effectiveness of different levels of features in alleviating the over-smoothing problem.
Compared with these baselines, F$^2$SAGE can achieve the best performance and higher MAD values by utilizing features in each block adaptively. It can further indicate the effectiveness of our method.


\begin{figure}[t]
\centering
	\includegraphics[width=0.8\linewidth]{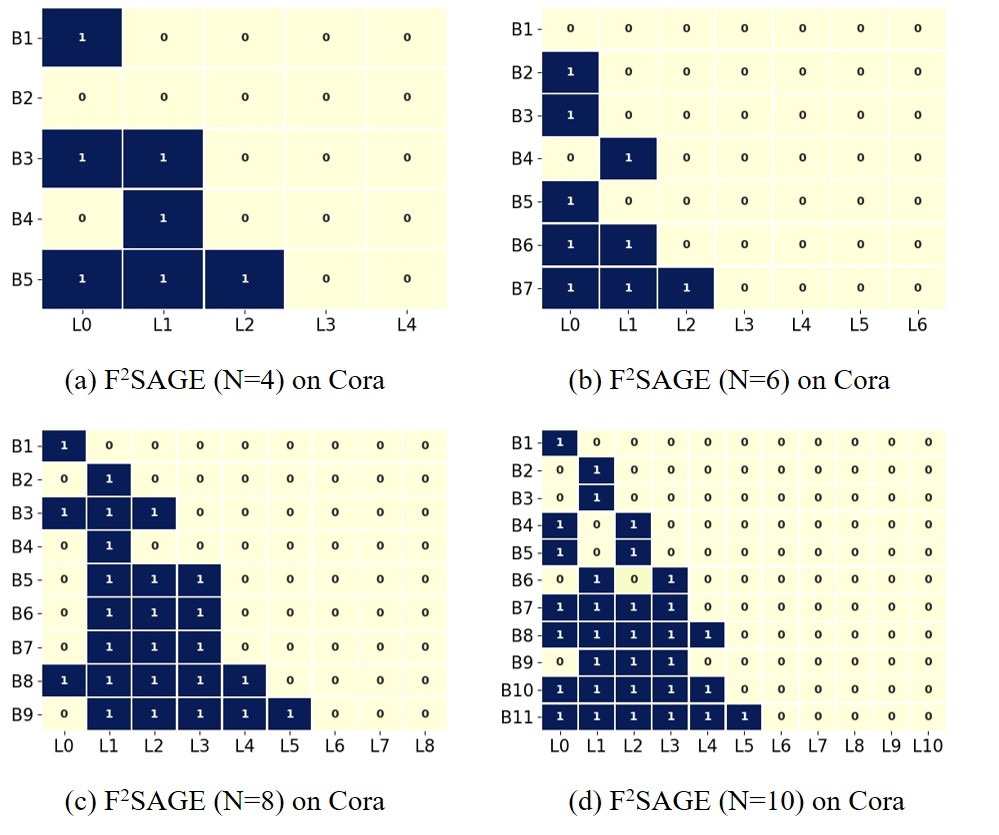}
	\centering
	\caption{We visualize the usage of different levels of features in F$^2$SAGE on the Cora dataset. $N$ is the number of SFA blocks. The cell $(Bi,Lj)=1$ represents the features of level $j$ are selected in Block $i$. }
	\label{fig-ranges-blocks}
\end{figure}

\noindent\textbf{Flexibility in obtaining the higher-level features.}
The multiple aggregation operations provide an adequate and independent feature extraction stage on local neighbors while are costly to obtain the higher-level information.
To make a comparison with this topology design manner, we visualize the utilization of different levels of features on the Cora dataset in Figure~\ref{fig-ranges-blocks}. If the features of level $j$ are selected in Block $i$, then the cell $(Bi, Lj)$ is denoted as 1, otherwise, as 0. 
%
We obtained the features of level 2 with four aggregation operations as shown in Figure~\ref{fig-ranges-blocks}(a) and obtained the features of level 5 with eight aggregation operations as shown in Figure~\ref{fig-ranges-blocks}(c). However, with the same number of aggregations, i.e., 8 and 4, PNA can only obtain the features of level 2 and level 1, respectively. 
%
Besides, on the Cora dataset, our method achieves higher performance than PNA with 35\% (0.39MB on F$^2$SAGE and 0.60MB on PNA) and 15\% (0.22MB on F$^2$GAT and 0.26MB on PNA) fewer parameters on the GraphSAGE and GAT, respectively.
Thus, our method is more efficient and flexible in obtaining the higher-level features.

From Figure~\ref{fig-ranges-blocks}, we can observe that the lower-level features, which are distinguishable and helpful for prediction~\cite{li2018deeper}, are more likely to be utilized in each block compared with higher-level ones. 
Besides, the over-smoothing problem is generated by stacking aggregation operations thus the features in higher-level become indistinguishable. This problem can be alleviated by selecting different levels of features in each block adaptively based on the proposed method.
As a summary, we evaluate the over-smoothing problem and visualize how different levels of features are utilized in our framework. The results demonstrate that our method can alleviate corresponding deficiencies of these two manners by utilizing features adaptively in each block. 
Combined with the performance gain and the top ranking in Table~\ref{tb-performance-agg}, the advantages of our method over the existing two topology design manners are significant and the aforementioned topology design target can be achieved with the proposed method.

\subsection{Ablation study}
In this section, we conduct ablation studies on the proposed framework.

\subsubsection{The Importance of Fusion Strategy}
In designing topologies, existing methods mainly focus on the links among operations~\cite{xie2019exploring,valsesia2020don}. By reviewing the extensive human-designed GNNs, we observe that the feature selection and fusion strategies result in the main difference of topology designs in GNNs. 
It indicates that the selection and fusion strategies are key components in designing the topology of GNNs, and the fusion strategies should be considered since they can improve the utilization of features in different levels~\cite{xu2018representation,zhao2020simplifying,zhao2021search}.
To evaluate the importance of fusion strategy, we learn to design topologies with fixed fusion operations instead. As shown in Table~\ref{tb-ablation-fusion}, with the three predefined fusion operations, the performance drop is evident on three variants, which demonstrates the importance of the fusion strategy. 
Therefore, designing the topology of GNNs with the selection and fusion operations is significant compared with the existing methods.
\begin{table}[]
	\caption{Performance comparisons of different fusion operations. The best results are highlighted and the second are underlined.}
	\centering
\footnotesize
\begin{tabular}{c|c|c|c}
	\hline
	Method        & Cora           & PubMed         & Physics        \\	\hline
F$^2$SAGE-SUM    & 84.73(0.63)               & 89.39(0.21)                 & 96.44(0.01)                 \\
F$^2$SAGE-MEAN   & 84.30(0.61)                & {\ul 89.58(0.22)}           & 96.42(0.03)                 \\
F$^2$SAGE-CONCAT & {\ul 86.07(0.45)}         & 89.31(0.19)                 & {\ul 96.69(0.01)}           \\ \hline
F$^2$SAGE        & \gray 87.72(0.26)         & \gray 89.73(0.26)           & \gray 96.72(0.01)                 \\ \hline
\end{tabular}
\label{tb-ablation-fusion}
\end{table}

%
%
%
%

\subsubsection{The Evaluation of the Improved Search Algorithm}
\label{sec-ablation-gap}

The efficiency of the differentiable search algorithm over others is significant and has been proved in existing methods~\cite{liu2018darts,zhao2021search,li2021one}. Therefore, we provide an improved differentiable search algorithm to design GNNs. 
In Table~\ref{tb-performance-agg}, SNAG and Random methods need to sample architectures and then train from scratch. In our experiments, they require 26.44 and 0.903 GPU hours to search GNNs on the Cora dataset. SANE and F$^2$GNN employ the differentiable search algorithm and the search cost are 0.007 and 0.028 GPU hours on the Cora dataset, respectively.

However, we observe that the optimization gap has a large influence on the feature selection operation due to the two opposite selection operations.  
As shown in Table~\ref{tb-gap}, we use the validation accuracy to quantify the optimization gap in the feature fusion framework. These accuracies are obtained at the end of the search and after architecture derivation without fine-tuning. 
We can observe that when the temperature is too large, i.e., 1 and 0.1, the validation accuracy gap of the supernet and the childnet is large when we search the feature selection and fusion operations. In these cases, the \texttt{ZERO} operation is selected for most features. 
However, in the F$^2$AGG method, which only searches aggregation operations based on the selection operation \texttt{IDENTITY} and fusion operation \texttt{SUM},
%
the accuracy gap is much smaller.
Therefore, considering the two opposite operations in the selection operation set, we use the small temperature of 0.001 instead.
The performance gains in Table~\ref{tb-performance-agg} indicate the effectiveness of the improved search algorithm in addressing the optimization gap in feature fusion and obtaining the expressive GNNs.  


\subsubsection{The Influence of the Number of SFA Blocks}
\label{sec-num-blocks}
\begin{figure}[t]
	\centering
	\includegraphics[width=0.9\linewidth]{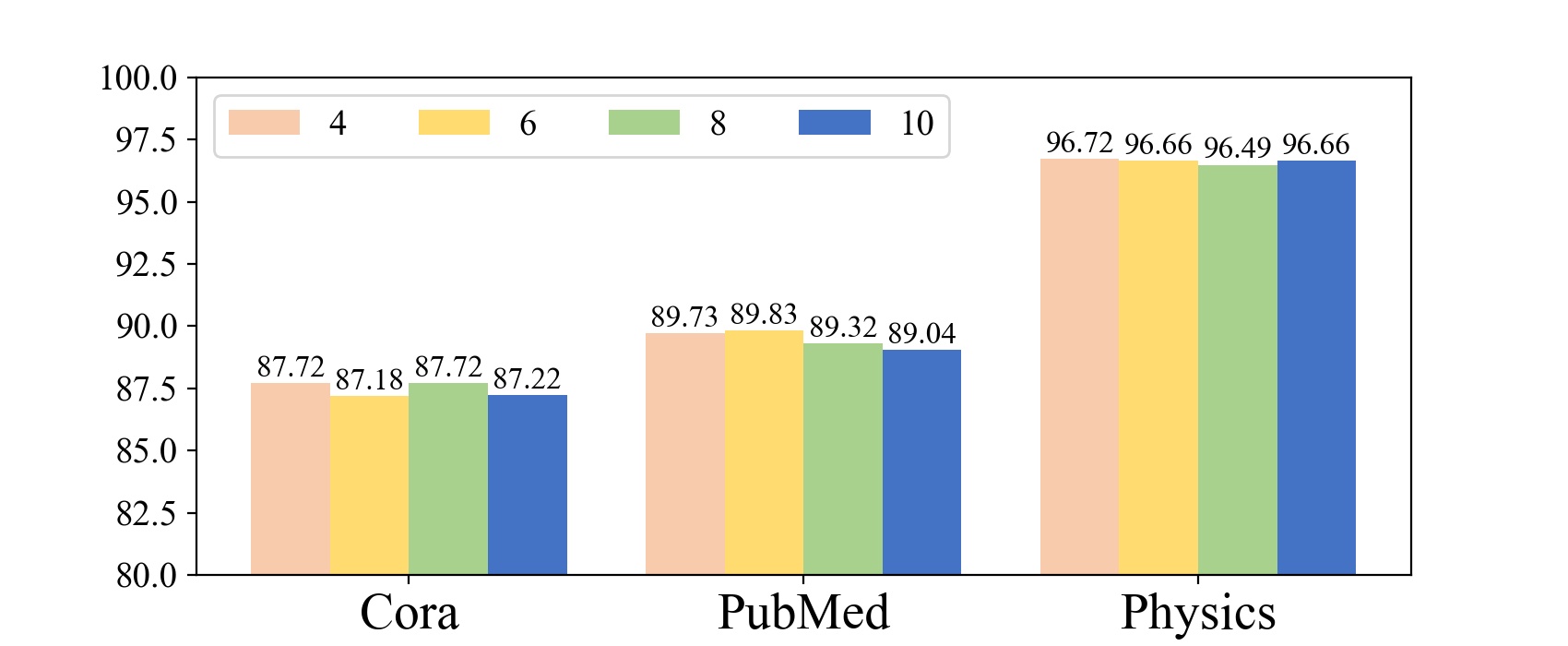}
	\caption{Performance comparisons of the different SFA block numbers on F$^2$SAGE. Different colors represent different block numbers.}
	\label{fig-ablation-blocks}
\end{figure}

In this paper, we only use four SFA blocks for example. Here we conduct experiments to show the influences of the number of SFA blocks. 
In Figure~\ref{fig-ablation-blocks}, F$^2$SAGE achieves a stable performance on the condition of different SFA blocks, yet the stacked baselines obtained the performance drop due to the over-smoothing problem as shown in Figure~\ref{fig-mad-layer}.
The increasing number of SFA blocks do not bring about the performance drop due to the adaptive utilization of different levels of features, which then demonstrates the strong ability of our method in alleviating the over-smoothing problem.

\begin{table}[t]
	\footnotesize
	\centering
	\caption{Comparisons of the validation accuracy in the supernet and the childnet on the Cora dataset.}
	\begin{tabular}{c|c|c|c|c}
		\hline
		\multirow{2}{*}{Temperature $\lambda$} & \multicolumn{2}{c|}{F$^2$SAGE} & \multicolumn{2}{c}{F$^2$AGG}                                \\ \cline{2-5} 
		& Supernet      & Childnet      & \multicolumn{1}{c|}{Supernet} & \multicolumn{1}{c}{Childnet} \\ \hline
		1                                      & 80.33          & 6.68          & 86.83         & 85.71         \\ \hline
		0.1                                    & 73.65          & 10.96         & 84.23         & 83.86         \\ \hline
		0.01                                   & 70.13          & 70.13         & 84.60         & 84.60         \\ \hline
		0.001                                  & 80.15          & 80.15         & 86.83         & 86.83         \\ \hline
	\end{tabular}
	\label{tb-gap}
\end{table}

\section{Conclusion and Future Work}

In this paper, we provide a novel feature fusion perspective in designing the GNN topology and propose a novel framework to unify the existing topology designs with feature selection and fusion strategies. In this way, designing the topology of GNNs is transformed into designing the feature selection and fusion strategies in the unified framework. 
To obtain an adaptive topology design, we develop a NAS method. 
To be specific, we provide a set of candidate selection and fusion operations in the search space and develop an improved differentiable search algorithm to address the obvious optimization gap in our method. 
Extensive experiments are conducted on the predefined and learnable aggregation operations on eight real-world datasets (five homophily and three heterophily).
The results demonstrate the effectiveness and versatility of the proposed method, and we can enjoy the benefits and alleviate the deficiencies of existing topology design manners, especially alleviating the over-smoothing problem, by utilizing different levels of features adaptively.

For future work, we will investigate the influence of different candidate operations and algorithms, and explore F$^2$GNN in large-scale graphs~\cite{hu2020ogb} and the heterophily datasets.

\begin{acks}
We thank Dr. Quanming Yao and Dr. Yongqi Zhang to improve the method and manuscript.
We also thank all anonymous reviewers for their constructive comments, which help us to improve the quality of this manuscript.
\end{acks}



\clearpage
\bibliographystyle{ACM-Reference-Format}
\bibliography{main} 

\clearpage
\appendix

\section{Optimization}
\label{appendix-alg}
\begin{algorithm}[t]
\small
	\caption{F$^2$GNN - Feature Fusion GNN}
	\begin{algorithmic}[1]
		\Require{Input graph $\mathcal{G} =(\mathcal{V}, \mathcal{E})$, the adjacency matrix $\bA$ and the feature matrix $\bH^{input}$. Training set $\mathcal{D}_{train}$ and validation set $\mathcal{D}_{valid}$, the epoch $T$ for search. }
		\Ensure{The searched architecture.}
		\State Random initialize the parameters $\bm{\alpha}$ and $\bW$.
		\For{$t=1$ to $T$}
		\State Calculate the operation weights \textbf{C} on top of $\bm{\alpha}$.
		\State $\bH^0 = \text{MLP}(\bH^{input})$ \label{alg-h0}
		\For {$i=1$ to $N+1$}
			\State $\bH^{s} = \{\}$
			\For {$j=0$ to $i-1$}
				\State $f_s^{ij}(\bH^j) = \sum\nolimits_{k=1}^{\left|\mathcal{O}_s\right|}  c_k^{ij}o_k^{ij}(\bH^j)=c_2^{ij}\bH^j$ $//$Selection
				\State $\bH^{s} = \bH^{s} \cup \{f_s^{ij}(\bH^j)\}$   
			\EndFor
			\State $\bH^{f} = f_f^i(\bH^s) = \sum\nolimits_{k=1}^{|\mathcal{O}_f|} c_k^io_k^i(\bH^s)$   $//$Fusion
			\If{$i=N+1$}
			\State $\bH^{output}=\text{MLP}(\bH^f)$
			\Else 
			\State $\bH^i = f_a^i(\bH^f, \bA)$  $//$Aggregation
			\EndIf
		\EndFor \label{alg-hout}
		\State Calculate the training loss $\mathcal{L}_{train}$ and update $\bW$.
		\State Calculate the operation weights \textbf{C} on top of $\bm{\alpha}$.
		\State Calculate the validation loss $\mathcal{L}_{valid}$ and update $\bm{\alpha}$.
		\EndFor
		\State Preserve the operation with the largest weight in each mixed operation.\\
		\Return The searched architecture.
	\end{algorithmic}
	\label{alg-f2gnn}
\end{algorithm}

%

In this paper, we optimize the supernet parameters $\bm{\alpha}$ and operation parameters $\bW$ with gradient descent as shown in Alg.~\ref{alg-f2gnn}. In the unified framework, the output $\bH^{output}$ can be calculated as shown in Line~\ref{alg-h0}-\ref{alg-hout}. 
The mixed selection operation $f_s^{ij}$ and fusion operation $f_f^i$ are given in Eq.~\eqref{eq-block-output} and Eq.~\eqref{eq-combination}, respectively. 
For the learnable aggregation operation, the mixed aggregation operationin in block $i$ can be calculated as $f_a^i(\bH, \bA)=\sum\nolimits_{k=1}^{|\mathcal{O}_a|} c_k^{i}o_k^i(\bH, \bA)$. For the predefined aggregation, $f_a^i(\bH, \bA)$ equals to the results of the predefined operation. 
After obtaining the output $\bH^{output}$, the cross-entropy loss $\mathcal{L}_{train}$ and $\mathcal{L}_{valid}$ can be generated and we can update the parameters with gradient descent as shown in Alg.~\ref{alg-f2gnn}.

\section{Details of experiments}

\subsection{Datasets}
\label{appendix-dataset}
Cora~\cite{sen2008collective}, DBLP~\cite{bojchevski2017deep} and PubMed~\cite{sen2008collective} are citation networks where each node represents a paper, and each edge represents the citation relation between two papers; 
Computers~\cite{mcauley2015image} is the Amazon co-purchase graph where nodes represent goods that are linked by an edge if these goods are frequently bought together; 
Physics~\cite{shchur2018pitfalls} is a co-authorship graph where nodes are authors who are connected by an edge if they co-author a paper;
Actor~\cite{pei2020geom} is a co-occurrence graph where each node correspond to an actor and each edge denotes co-occurrence of these two actors on the same Wikipedia page; 
Wisconsin~\cite{pei2020geom} is the hyperlinked web page graph where nodes represent web pages and edges are hyperlinks between them;
In Flickr~\cite{zeng2019graphsaint}, nodes represent images and edges represent two images that share some common properties (e.g.,
same geographic location, same gallery, comments by the same user, etc.).

\subsection{Baselines}
\label{appendix-baseline}
To make a fair comparison with F$^2$SAGE and F$^2$GAT, we provide nine GNNs constructed with the predefined aggregation operation and different topology designs. For two stacking baselines, RES and JK methods, the topology designs of these methods are shown in Figure~3.
The topology design of the DENSE method is constructed based on DenseGCN~\cite{li2019deepgcns}. To be more specific, the aggregation operations are densely connected and the fusion operation \texttt{CONCAT} is utilized in each layer.

For GCNII, the aggregation operation can be represented by $\bH^{l+1}=\sigma(((1-\alpha_l)\tilde{\textbf{P}}\bH^l + \alpha_l\bH^0)((1-\beta_l)\textbf{I}_n + \beta_l\bW^l)$. $\alpha_l$ and $\beta_l$ are the hyperparameters,  $\tilde{\textbf{P}}=\tilde{\textbf{D}}^{-1/2}\tilde{\textbf{A}}\tilde{\textbf{D}}^{-1/2}$ is the graph convolution matrix with the renormalization trick. 
For simplicity, we ignore the feature transformation thus the feature fusion strategy in GCNII can be represented as $\bH^{l+1} = \alpha \bH^0 + (1-\alpha)f_a(\bH^l)$ where $f_a$ is the aggregation operation. The illustration is shown in Figure 3(d). 
The GNNII baseline in our experiment is constructed on top of Figure 3(d) where we add the features of $\bH^0$ and $\bH^{i-1}$ in Block $i$, then an aggregation operation or MLP followed behind.

In Figure 3(e), one layer PNA with four aggregation operations is provided. The PNA baseline in our experiment is constructed by stacking two layers with eight aggregation operations. In Figure 3(f), one layer MixHop with three aggregation operations is provided and the MixHop baseline in our experiment is constructed by stacking two layers.

We provide the performance comparisons of the GCNII, PNA and MixHop baselines used in our experiment and used in PyG~\footnote{https://github.com/pyg-team/pytorch\_geometric/tree/master/examples}. As shown in Table~\ref{tb-performance-appendix}, our baselines can achieve considerable performance on top of the unified framework with the same evaluation stage which will be introduced in the following.

\begin{figure}[t]
	\centering
	\includegraphics[width=0.5\linewidth]{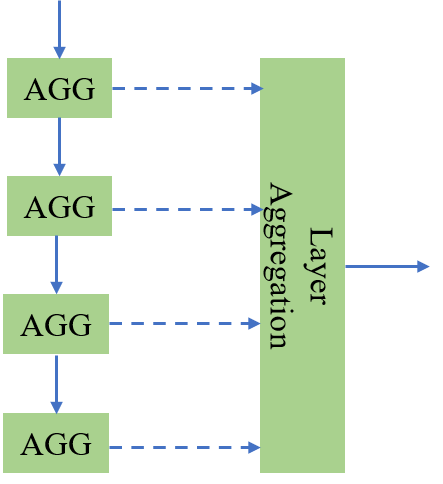}
	\caption{The 4-layer GNN backbone used in SANE and SNAG. The ``AGG'' rectangle represents the aggregation operation in each layer. The dashed lines represent the learnable connections between the intermediate layers and the layer aggregation. The ``Layer aggregation'' rectangle represents the fusion operation used to integrate these selected features.}
	\label{fig-sane-framework}
\end{figure}

Compared with F$^2$GNN, we provide the SNAG and SANE baselines. As shown in Figure~\ref{fig-sane-framework}, they focus on designing the aggregation operations in each layer, connections and layer aggregations in the output block. 
The search spaces are shown in Table~\ref{tb-space-sane}.

\begin{table}[]
	\caption{The search spaces in SANE and SNAG.}
	\footnotesize
	\label{tb-space-sane}
	\begin{tabular}{C{65pt}|L{65pt}|L{65pt}}
		\toprule
		& SANE & SNAG \\ \hline
		Aggregation & \texttt{SAGE-SUM}, \texttt{SAGE-MEAN}, \texttt{SAGE-MAX}, \texttt{GCN}, 
		\texttt{GAT}, \texttt{GAT-SYM}, \texttt{GAT-COS}, \texttt{GAT-LINEAR}, \texttt{GAT-GEN-LINEAR}, \texttt{GIN}, \texttt{GeniePath}      & \texttt{GCN}, \texttt{SAGE-SUM}, \texttt{SAGE-MEAN}, \texttt{SAGE-MAX}, \texttt{MLP},  
		\texttt{GAT}, \texttt{GAT-SYM}, \texttt{GAT-COS}, \texttt{GAT-LINEAR}, \texttt{GAT-GEN-LINEAR}   \\ \hline
		Skip-connection    & \texttt{ZERO}, \texttt{IDENTITY}      & \texttt{ZERO}, \texttt{IDENTITY}      \\  \hline
		Layer aggregation      & \texttt{CONCAT}, \texttt{MAX}, \texttt{LSTM}     &  \texttt{CONCAT}, \texttt{MAX}, \texttt{LSTM}  \\ \bottomrule   
	\end{tabular}
\end{table}

\begin{table*}[]
	\footnotesize
	\centering
	\caption{Performance comparisons of the baselines used in our experiment and used in PyG. We report the average test accuracy and the standard deviation with 10 splits. ``L2'' and ``L4'' mean the number of layers of the base GNN architecture, respectively. ``OOM'' means out of memory.}
	\label{tb-performance-appendix}
	\begin{tabular}{c|c|cccccccc}
		\hline
		& Topology    & Cora        & DBLP        & PubMed      & Computer    & Physics     & Actor       & Wisconsin   & Flickr      \\ \hline
		\multirow{3}{*}{SAGE} & GNNII (L4)  & 85.83(0.42) & 84.46(0.45) & 89.21(0.24) & 91.38(0.27) & 96.45(0.15) & 35.70(1.11)  & 81.57(4.13) & 52.24(0.29) \\ 
		& PNA (L2)    & 84.29(0.67) & 82.76(0.42) & 89.25(0.26) & 90.67(0.42) & 96.32(0.10)  & 33.89(2.68) & 75.29(6.46) & 52.09(0.73) \\ 
		& MixHop (L2) & 84.81(0.95) & 82.65(0.65) & 89.25(0.28) & 88.56(1.61) & 96.11(0.17) & 35.19(0.62) & 81.57(2.51) & 51.75(0.59) \\ \hline
		\multirow{3}{*}{GAT}  
		& GNNII (L4)  & 85.40(1.06)  & 83.83(0.33) & 88.44(0.25) & 91.91(0.11) & 96.14(0.15) & 30.29(0.78) & 55.29(6.25) & 53.03(0.29) \\ 
		& PNA (L2)    & 85.06(0.72) & 83.46(0.47) & 86.81(0.47) & 90.84(0.24) & 95.85(0.18) & 28.14(1.99) & 47.65(5.12) & 54.02(0.33) \\ 
		& MixHop (L2) & 85.38(1.04) & 82.50(0.34)  & 88.91(0.19) & 91.27(0.37) & 96.46(0.21) & 35.70(0.90)   & 81.57(4.40)  & 53.67(0.30)  \\ \hline
		\multirow{3}{*}{PyG}  & GCNII (L4)  & 83.27(1.14) & 82.78(0.56) & 89.03(0.22) & 42.08(3.07) & 96.08(0.07) & 32.16(2.83) & 67.25(7.89) & 42.91(0.82) \\ 
		& PNA (L2)    & 86.99(0.57) & 84.05(0.18) & 88.99(0.15) & 91.61(0.11) & 96.60(0.11)  & 33.43(1.35) & 68.24(6.00)  & 52.18(0.17) \\ 
		& MixHop (L2) & 86.51(0.55) & 82.24(0.50)  & 88.58(0.38) & 90.57(0.33) & 96.53(0.14) & 36.89(0.77) & 78.04(4.19) & OOM         \\ \hline
	\end{tabular}
\end{table*}

\subsection{Implementation details}
\label{appendix-implementation}
All models are implemented with Pytorch \cite{paszke2019pytorch} on a GPU 2080Ti (Memory: 12GB, Cuda version: 10.2).
Thus, for consistent comparisons of baseline models, we use the implementation of all GNN baselines by the popular GNN library: Pytorch Geometric (PyG) (version 1.6.1)~\cite{Fey/Lenssen/2019}~\footnote{https://github.com/pyg-team/pytorch\_geometric}.

For SNAG~\cite{zhao2020simplifying}~\footnote{https://github.com/AutoML-Research/SNAG}, an RL based method to design the aggregation operations in each layer, connections and layer aggregations in the output block, we use the 4-layer backbone to make a fair comparison with the proposed methods. In the search process, we set the search epoch to 500.
In each epoch, we sample 10 architectures and use the validation accuracy to update the controller parameters. After training finished, we sample 5 candidates with the controller. 

For SANE~\cite{zhao2021search}~\footnote{https://github.com/AutoML-Research/SNAE}, a differentiable method to design aggregation operations in each layer, connections and layer aggregations in the output block, the 4-layer backbone is utilized. In the search process, we set the search epoch to 30. One candidate GNN can be obtained after the search process. Repeat 5 times with different seeds, we can get 5 candidates. 

For each random baseline in Table~\ref{tb-performance-agg}, we randomly sample 100 GNNs from the designed search space and train these methods from scratch. 1 candidate GNN is derived based on the validation accuracy.

For our method,  we set the search epoch to 400 and $\lambda$ to 0.001. One candidate GNN can be obtained after the search process. Repeat 5 times with different seeds, we can get 5 candidates. 

The searched GNNs and all human-designed baselines are finetuned individually with the hyperparameters shown in Table~\ref{tb-hypers}.
Each method owns 30 hyper steps. In each hyper step, a set of hyperparameters will be sampled from Table~\ref{tb-hypers} based on Hyperopt~\footnote{https://github.com/hyperopt/hyperopt}, and then generate final performance on the split data. 
We choose the hyperparameters for each candidate with the validation accuracy, and then select the candidate for SNAG, SANE and the proposed method with the validation accuracy.

After that, we report the final test accuracy and the standard deviations based on 10 repeat runs.

\begin{table}[]
	\centering
	\small
	\caption{Hyperparameters we used in this paper.} 
	\begin{tabular}{l|l}
		\hline
		hyperparameter      & Operation \\ \hline
		Embedding size      & 16, 32, 64, 128, 256, 512          \\ \hline
		Learning rate       &  $[0.001, 0.01]$         \\ \hline
		Dropout rate        & 0, 0.1, 0.2,$\cdots$,0.9           \\ \hline
		Weight decay        & $[0.0001, 0.001]$          \\ \hline
		Optimizer           &  Adam, AdaGrad         \\ \hline
		Activation function & Relu, ELU \\  \hline
	\end{tabular}
	\label{tb-hypers}
\end{table}

\subsection{Searched Topologies}
\label{appendix-searched}
\begin{figure}[t]
\footnotesize
\centering
	\subfigure[F$^2$GNN on Cora]{
		\begin{minipage}[t]{0.45\linewidth}
			\centering
			\includegraphics[width=1.0\linewidth]{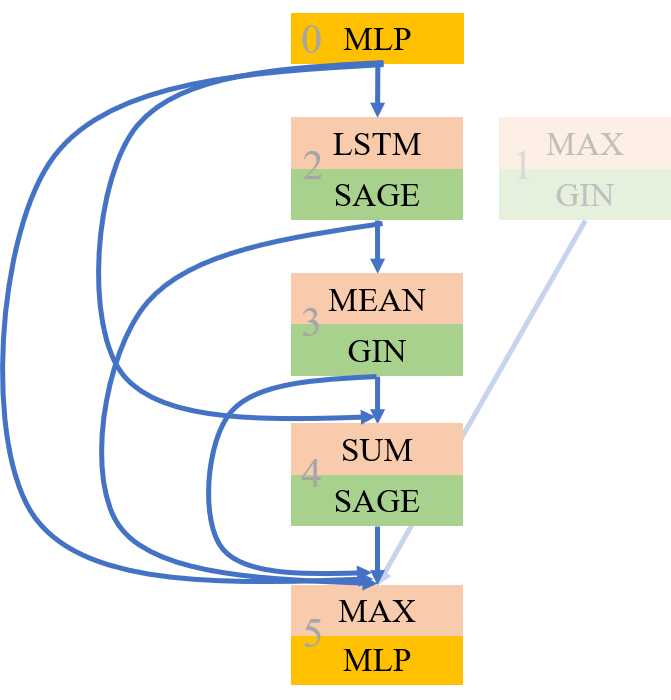}
		\end{minipage}%
	}%
	\subfigure[F$^2$GNN on Phisics]{
		\begin{minipage}[t]{0.45\linewidth}
			\centering
			\includegraphics[width=1.0\linewidth]{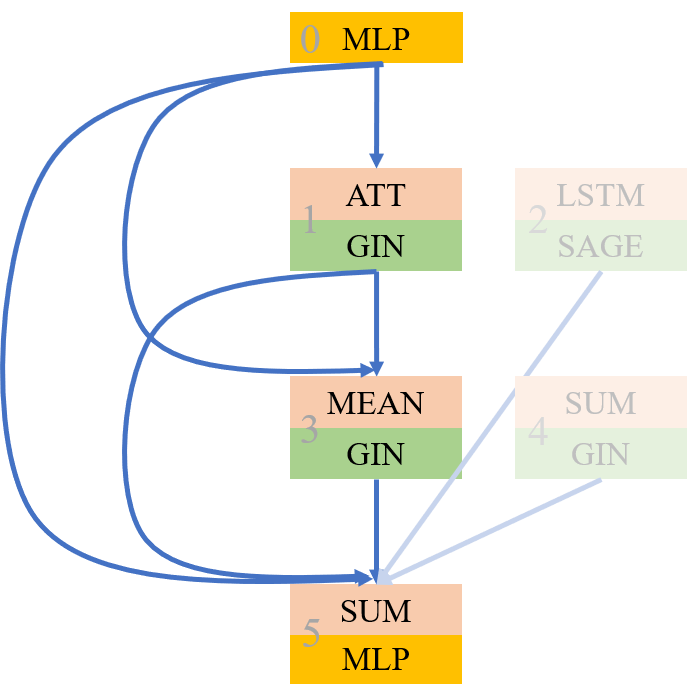}
		\end{minipage}%
	}%
	\\
	\subfigure[F$^2$GNN on Actor]{
		\begin{minipage}[t]{0.45\linewidth}
			\centering
			\includegraphics[width=1.0\linewidth]{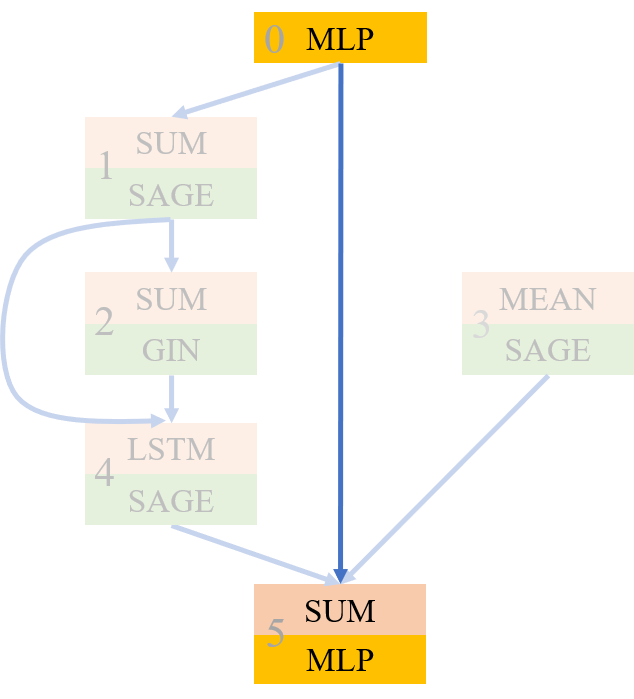}
		\end{minipage}%
	}%
	\caption{The searched topologies on Cora, Physics and Actor datasets with F$^2$GNN.}
	\label{fig-searched-appendix}
\end{figure}

For sake of space, we only show the searched topologies of F$^2$SAGE and F$^2$GAT. The results of F$^2$GNN are shown in Figure~\ref{fig-searched-appendix}.

\begin{figure}[t]
	\centering
	\includegraphics[width=0.9\linewidth]{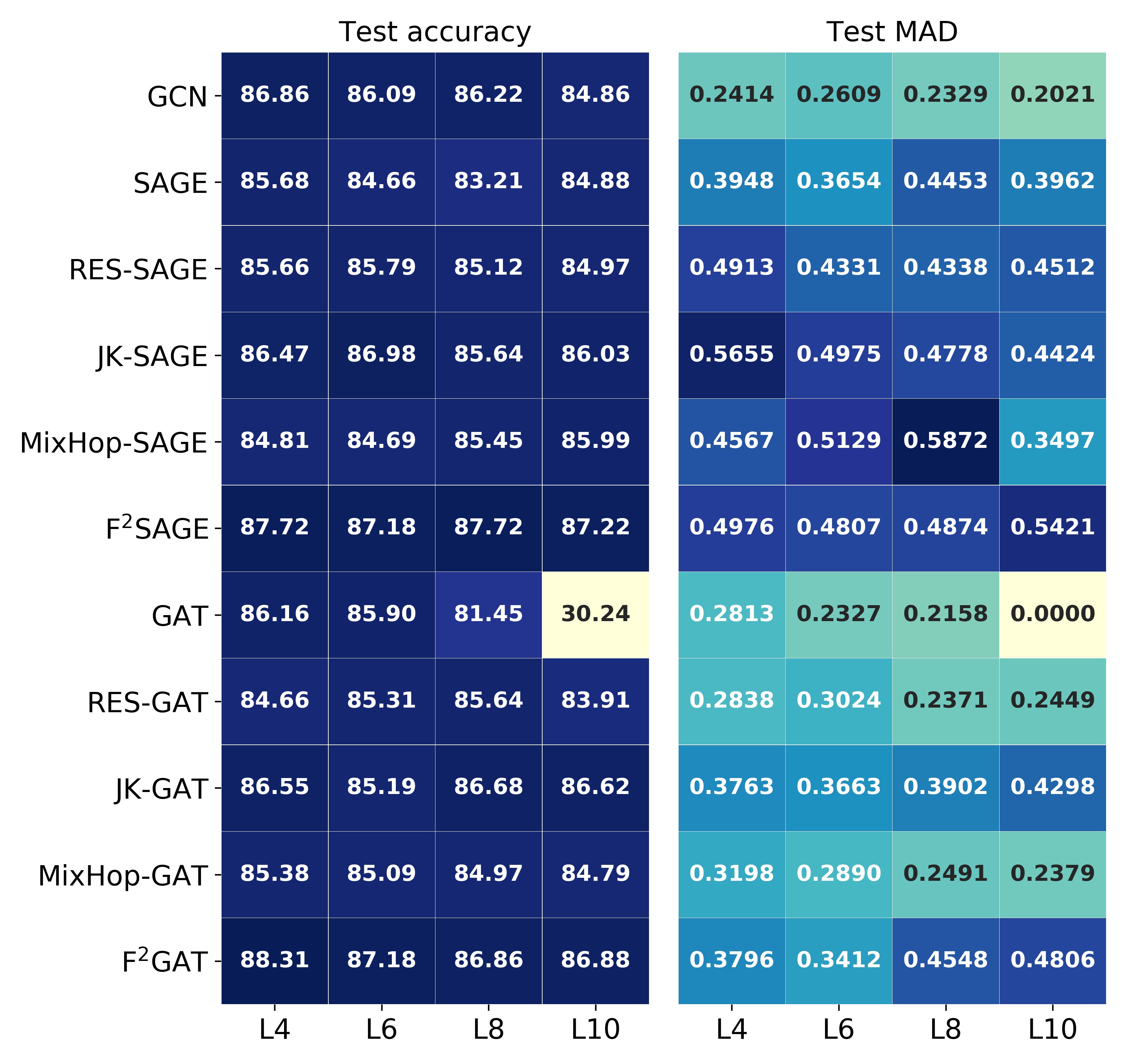}
	\caption{Comparisons of test accuracy and the MAD value on the Cora dataset. Predefined aggregation operations are the GraphSAGE and GAT. ``L4'' represents the 4 layer baseline ($N=4$ in our method), and so on. Darker colors mean larger values.}
	\label{fig-mad-layer-full}
\end{figure}

\subsection{Heterophily results}
\label{appendix-hetero}
\subsubsection{Datasets}
As shown in Table~\ref{tb-dataset_hetero}, Actor~\cite{pei2020geom} is a graph representing actor co-occurrence in Wikipedia pages, Texas, Wisconsin and Cornell are the hyperlinked web pages of various universities provided by~\cite{pei2020geom}, Chameleon and Squirrel are the subgraphs of web pages in Wikipedia discussing the corresponding topics provided by~\cite{rozemberczki2021multi}.

\subsubsection{Performance comparisons}
On the heterophily datasets where the connected nodes are from different classes in most cases, the general aggregation operations cannot capture enough information from the graph structure and the MLP can even outperform several existing baselines as shown in H$_2$GCN\cite{zhu2020beyond}. 
It proved that the features generated in the intermediate layers help to increase the representation power of the model under heterophily, and the performance comparisons results in Table~\ref{tb-performance-heterophily} can demonstrate the effectiveness of this method. Compared with H$_2$GCN which selected features in the output block, our method utilizes different levels of features in each block which has a more flexible feature utilization stage than H$_2$GCN. The top ranking we achieve can demonstrate the effectiveness of the feature fusion method in heterophily settings.

\begin{table}[]
	\centering
	\small
	\caption{Statistics of the six heterophily datasets.}

\begin{tabular}{c|c|c|c|c|c}
	\hline
	Datasets                               & \#Nodes & \#Edges & \#Features & \#Classes & h    \\ \hline
	Texas~\cite{pei2020geom}               & 183     & 309     & 1703       & 5         & 0.11 \\ \hline
	Wisconsin~\cite{pei2020geom}           & 251     & 466     & 1,703      & 5         & 0.21 \\ \hline
	Actor~\cite{pei2020geom}               & 7,600   & 30,019  & 932        & 5         & 0.22 \\ \hline
	Squirrel~\cite{pei2020geom}            & 5,201   & 198,493 & 2,089       & 5         & 0.22 \\ \hline
	Chameleon~\cite{rozemberczki2021multi} & 2,277   & 31,421  & 2,325      & 5         & 0.23 \\ \hline
	Cornell~\cite{pei2020geom}             & 183     & 280     & 1,703      & 5         & 0.3  \\ \hline
\end{tabular}
	\label{tb-dataset_hetero}
\end{table}

\begin{table*}[]
	\small
	\caption{Performance comparisons on the heterophily datasets. The ``$\ast$'' results of all these methods are obtained from~\cite{zhu2020beyond}. The best results in this table are highlighted in gray, and the second results are underlined. We provide the average accuracy rank as well and highlight the Top-3 methods in this table.}
	\begin{tabular}{c|c|c|c|c|c|c|c}
		\hline
		Method             & Texas                         & Wisconsin                     & Actor                         & Squirrel                       & Chameleon                     & Cornell                       & Avg. Rank  \\ \hline
		F$^2$SAGE          & \underline{$0.8405^{0.0409}$} & $0.8588^{0.0192}$             & $0.3661^{0.0100}$             & $0.3604^{0.0148}$              & $0.5864^{0.0186}$             & $0.8324^{0.0637}$             & 6.5        \\
		F$^2$GCN           & $0.8378^{0.0527}$             & $0.8608^{0.0571}$             & \underline{$0.3701^{0.0101}$} & $0.4181^{0.0157}$              & $0.6331^{0.0086}$             & \gray $0.8352^{0.0656}$       & \gray 3.5  \\
		F$^2$GAT           & $0.8270^{0.0595}$             & \underline{$0.8706^{0.0413}$} & $0.3665^{0.0113}$             & \gray $0.4732^{0.0243}$        & \gray $0.6781^{0.0205}$       & \underline{$0.8351^{0.0670}$} & \gray 2.5  \\
		F$^2$GNN           & $0.8297^{0.0514}$             & \gray $0.8824^{0.0372}$       & \gray $0.3708^{0.0100}$       & $0.3658^{0.0167}$              & \underline{$0.6757^{0.0216}$} & $0.8270^{0.0757}$             & \gray 4.17 \\ \hline
		H$_2$GCN-1$\ast$   & \gray $0.8486^{0.0677}$       & $0.8667^{0.0469}$             & $0.3586^{0.0103}$             & $0.3642^{0.0189}$              & $0.5711^{0.0158}$             & $0.8216^{0.0480}$             & 6.67       \\
		H$_2$GCN-2$\ast$   & $0.8216^{0.0528}$             & $0.8588^{0.0422}$             & $0.3562^{0.0130}$             & $0.3790^{0.0202}$              & $0.5939^{0.0198}$             & $0.8216^{0.0600}$             & 7.67       \\
		GraphSAGE$\ast$    & $0.8243^{0.0614}$             & $0.8118^{0.0556}$             & $0.3423^{0.0099}$             & $0.4161^{0.0074}$              & $0.5873^{0.0168}$             & $0.7595^{0.0501}$             & 8.5        \\
		GCN-Cheby$\ast$    & $0.7730^{0.0407}$             & $0.7941^{0.0446}$             & $0.3411^{0.0109}$             & $0.4386^{0.0164}$              & $0.5524^{0.0276}$             & $0.7432^{0.0746}$             & 10.5       \\
		MixHop$\ast$       & $0.7784^{0.0773}$             & $0.7588^{0.0490}$             & $0.3222^{0.0234}$             & $0.4380^{0.0148}$              & $0.6050^{0.0253}$             & $0.7351^{0.0634}$             & 9.83       \\ \hline
		GraphSAGE+JK$\ast$ & $0.8378^{0.0221}$             & $0.8196^{0.0496}$             & $0.3428^{0.0101}$             & $0.4085^{0.0129}$              & $0.5811^{0.0197}$             & $0.7568^{0.0403}$             & 8.17       \\
		Cheby+JK$\ast$     & $0.7838^{0.0637}$             & $0.8255^{0.0457}$             & $0.3514^{0.0137}$             & \underline{$0.4503^{0.0173}$ } & $0.6379^{0.0227}$             & $0.7459^{0.0787}$             & 6.83       \\
		GCN+JK$\ast$       & $0.6649^{0.0664}$             & $0.7431^{0.0643}$             & $0.3418^{0.0085}$             & $0.4045^{0.0161}$              & $0.6342^{0.0200}$             & $0.6459^{0.0868}$             & 10.3       \\ \hline
		GCN$\ast$          & $0.5946^{0.0525}$             & $0.5980^{0.0699}$             & $0.3026^{0.0079}$             & $0.3689^{0.0134}$              & $0.5982^{0.0258}$             & $0.5703^{0.0467}$             & 13.3       \\
		GAT$\ast$          & $0.5838^{0.0445}$             & $0.5529^{0.0871}$             & $0.2628^{0.0173}$             & $0.3062^{0.0211}$              & $0.5469^{0.0195}$             & $0.5892^{0.0195}$             & 15.5       \\
		GEOM-GCN$\ast$     & 0.6537                        & 0.6412                        & 0.3163                        & 0.3814                         & 0.609                         & 0.6081                        & 11.83      \\ \hline
		MLP$\ast$          & $0.8189^{0.0478}$             & $0.8529^{0.0361}$             & $0.3576^{0.0098}$             & $0.2968^{0.0181}$              & $0.4636^{0.0252}$             & $0.8108^{0.0637}$             & 10.17      \\ \hline
	\end{tabular}
	\label{tb-performance-heterophily}
\end{table*}

\subsection{Alleviating the over-smoothing problem}
\label{appendix-oversmoothing}
For sake of space, we only provide 4 baselines in Figure~\ref{fig-mad-layer}, and the complete results are shown in Figure~\ref{fig-mad-layer-full}.

%
%
%
%

\end{document}